\pdfoutput = 1
\documentclass[11pt]{article}
\usepackage{fullpage}
\usepackage[ruled,linesnumbered]{algorithm2e}
\usepackage{waingarten}
\usepackage{lineno}
\usepackage[pagebackref]{hyperref}  
\usepackage{cleveref}
 \newtheorem{lemma}{Lemma}[section]
  \newtheorem{theorem}[lemma]{Theorem}
\newtheorem{claim}[lemma]{Claim}

\newtheorem{corollary}[lemma]{Corollary}

\newtheorem{definition}[lemma]{Definition}
\usepackage{enumitem}
\usepackage{booktabs} 
\usepackage[colorinlistoftodos,prependcaption,textsize=tiny]{todonotes}

\usepackage[utf8]{inputenc} 
\usepackage[T1]{fontenc}    
\usepackage{hyperref}       
\usepackage{url}            
\usepackage{booktabs}       
\usepackage{amsfonts}       
\usepackage{nicefrac}       
\usepackage{microtype}

\usepackage[ruled,linesnumbered]{algorithm2e}
\usepackage{amsmath}
\usepackage{amsthm}
\usepackage{graphicx}
\usepackage{cleveref}
\usepackage{enumitem}
\usepackage{authblk}

\newcommand{\nnz}{\mathsf{nnz}}
\newcommand{\sps}[1]{^{(#1)}}

\SetKwFunction{fsum}{Sum}
\SetKwFunction{find}{Find}
\SetKwFunction{update}{Update}
\SetKwFunction{initialize}{Initialize}
\SetKwFunction{get}{Get}
\SetKw{And}{and}
\SetKwFunction{name}{PRONE}

\usepackage{wrapfig}
\usepackage{mathtools}
\newcommand{\erclogowrapped}[1]{
\setlength\intextsep{0pt}
\begin{wrapfigure}[3]{r}{#1*\real{1.1}}
\includegraphics[width=#1]{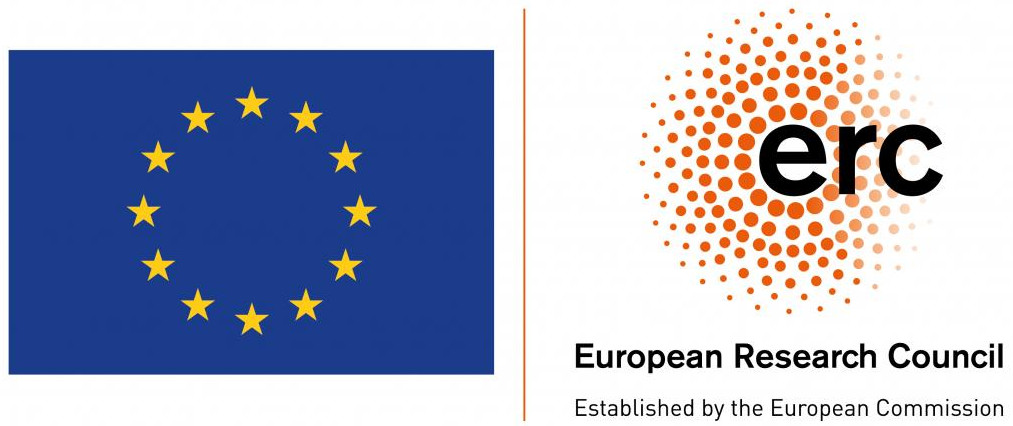}
\end{wrapfigure}
}

\title{Simple, Scalable and Effective Clustering via One-Dimensional Projections}
\author[1]{Moses Charikar}
\author[2]{Monika Henzinger}
\author[3]{Lunjia Hu}
\author[4]{Maximilian V\"otsch}
\author[5]{Erik Waingarten}
\affil[1,3]{Stanford University}
\affil[2]{Institute of Science and Technology Austria (ISTA)}
\affil[4]{Faculty of Computer Science, Doctoral School of Computer Science DoCS Vienna, University of Vienna}
\affil[5]{Department of Computer and Information Sciences, University of Pennsylvania}

\allowdisplaybreaks
\begin{document}

\maketitle

\begin{abstract}
  Clustering is a fundamental problem in unsupervised machine learning with many applications in data analysis. Popular clustering algorithms such as Lloyd's algorithm and $k$-means++ can take $\Omega(ndk)$ time when clustering $n$ points in a $d$-dimensional space (represented by an $n\times d$ matrix $X$) into $k$ clusters. 
  In applications with moderate to large $k$, the multiplicative $k$ factor can become very expensive. We introduce a simple randomized clustering algorithm that provably runs in expected time $O(\nnz(X) + n\log n)$ for arbitrary $k$. 
  Here $\nnz(X)$ is the total number of non-zero entries in the input dataset $X$, which is upper bounded by $nd$ and can be significantly smaller for sparse datasets. We prove that our algorithm achieves approximation ratio $\smash{\widetilde{O}(k^4)}$ on any input dataset for the $k$-means objective. We also believe that our theoretical analysis is of independent interest, as we show that the approximation ratio of a $k$-means algorithm is approximately preserved under a class of projections and that $k$-means++ seeding can be implemented in expected $O(n \log n)$ time in one dimension. Finally, we show experimentally that our clustering algorithm gives a new tradeoff between running time and cluster quality compared to previous state-of-the-art methods for these tasks.
\end{abstract}

\section{Introduction}
Clustering is an essential and powerful tool for data analysis with broad applications in computer vision and computational biology, and it is one of the fundamental problems in unsupervised machine learning. In large-scale applications, datasets often contain billions of high-dimensional points. Grouping similar data points into clusters is crucial for understanding and organizing datasets. Because of its practical importance, the problem of designing efficient and effective clustering algorithms has attracted the attention of numerous researchers for many decades.

One of the most popular algorithms for the $k$-means clustering problem is Lloyd's algorithm~\cite{1056489}, which seeks to locate $k$ centers in the space that minimize the sum of squared distances from the points of the dataset to their closest center (we call this the ``$k$-means cost''). While finding the centers minimizing the objective is NP-hard~\cite{ADHP09}, in practice we can find high-quality sets of centers using Lloyd's iterative algorithm.
Lloyd's algorithm maintains a set of $k$ centers. It iteratively updates them by assigning points to one of $k$ clusters (according to their closest center), then redefining the center as the points' center of mass.
It needs a good initial set of centers to obtain a high-quality clustering and fast convergence. In practice, the $k$-means++ algorithm \cite{MR2485254}, a randomized \emph{seeding} procedure, is used to choose the initial $k$ centers. $k$-means++ achieves an $O(\log k)$-approximation ratio in expectation, upon which each iteration of Lloyd's algorithm improves.\footnote{Approximation is with respect to the $k$-means cost. A $c$-approximation has $k$-means cost, which is at most $c$ times larger than the optimal $k$-means cost.} Beyond their effectiveness, these algorithms are simple to describe and implement, contributing to their popularity.

The downside of these algorithms is that they do not scale well to massive datasets. A standard implementation of an iteration of Lloyd's algorithm needs to calculate the distance to each center for each point in the dataset, leading to a $\Theta(ndk)$ running time. Similarly, the standard implementation of the $k$-means++ seeding procedure produces $k$ samples from the so-called $D^2$ distribution (see \Cref{sec:prelims} for details). Maintaining the distribution requires making a pass over the entire dataset after choosing each sample.  
Generating $k$ centers leads to a $\Theta(ndk)$ running time. Even for moderate values of $k$, making $k$ passes over the entire dataset can be prohibitively expensive. 

One particularly relevant application of large-scale $k$-means clustering is in approximate nearest neighbor search~\cite{simhadri2022results} (for example, in product quantization~\cite{jegou2010product} and building inverted file indices~\cite{FAISSMANUAL}). There, $k$-means clustering is used to compress entire datasets by mapping vectors to their nearest centers, leading to billion-scale clustering problems with large $k$ (on the order of hundreds or thousands). Other applications on large datasets requiring a large number of centers may be spam filtering \cite{10.1145/1811099.1811090, 10.1007/978-3-319-30303-1_12}, near-duplicate detection \cite{10.14778/1687627.1687771}, and compression or reconciliation tasks \cite{rostami2018interactive}.
New algorithmic ideas are needed for these massive scales, and this motivates the following challenge:
\begin{quote}
Can we design a simple, practical algorithm for $k$-means that runs in time roughly $O(nd)$, independent of $k$, and produces high-quality clusters?
\end{quote}

Given its importance in theory and practice, a significant amount of effort has been devoted to algorithms for fast $k$-means clustering. We summarize a few of the approaches below with the pros and cons of each so that we may highlight our work's position within the literature:
\begin{enumerate}[label=\Alph*.]
    \item \textbf{Standard $k$-means++}: This is our standard benchmark. 
    {\bf Plus:} Guaranteed to be an $O(\log k)$-approximation~\cite{MR2485254}; outputs centers, as well as the assignments of dataset points to centers. 
    {\bf Minus:} The running time is $O(ndk)$, which is prohibitively expensive in large-scale applications. 
    
    \item \textbf{Using Approximate Nearest Neighbor Search}: One may implement $k$-means++ faster using techniques from approximate nearest neighbor search (instead of a brute force search each iteration). 
    {\bf Plus:} The algorithms with provable guarantees, like \cite{NEURIPS2020_babcff88}, obtain an $\smash{O_{\varepsilon}(\log k)}$-approximation. 
    {\bf Minus:} The running time is $\smash{\widetilde O(nd + (n \log (\Delta))^{1 + \varepsilon})}$, depending on a dataset dependent parameter $\Delta$, the ratio between the maximum and minimum distances between input points. The techniques are algorithmically sophisticated and incur extra poly-logarithmic factors (hidden in $\smash{\widetilde{O}(\cdot)}$), making the implementation significantly more complicated.
    
    \item \textbf{Approximating the $D^2$-Distribution}: Algorithms that speed up the seeding procedure for Lloyd's algorithm or generate fast coresets (we expand on this below) have been proposed in \cite{Bachem_Lucic_Hassani_Krause_2016,NIPS2016_d67d8ab4,10.1145/3219819.3219973}. 
    {\bf Plus:} These algorithms are fast, making only one pass over the dataset in time $O(nd)$. (For \cite{Bachem_Lucic_Hassani_Krause_2016,NIPS2016_d67d8ab4}, there is an additional additive $O(k^2d)$ term in the running time). 
    {\bf Minus:} The approximation guarantees are qualitatively weaker than the approximation of $k$-means clustering. They incur an additional additive approximation error that grows with the entire dataset's variance (which can lead to an arbitrarily large error; see Section~\ref{sec:experiments}).
    These algorithms output a set of $k$ centers but not the cluster assignments. Naively producing the assignments would take time $O(ndk)$.\footnote{One may use approximate nearest neighbor search techniques to improve on the $O(ndk)$ running time. However, as discussed above, approximate nearest neighbor search adds a significant layer of complexity (and approximation).}
\end{enumerate}

\paragraph{Coresets.} 
At a high level, coresets are a dataset-reduction mechanism. A large dataset $X$ of $n$ points in $\R^d$ is distilled into a significantly smaller (weighted) dataset $Y$ of $m$ points in $\R^d$, called a ``coreset'' which serves as a good proxy for $X$, i.e., the clustering cost of any $k$ centers on $Y$ is approximately the cost of the same centers on $X$.
We point the reader to~\cite{bachem2017practical,F20} for a recent survey on coresets. Importantly, coreset constructions (with provable multiplicative-approximation guarantees) require an initial approximate clustering of the original dataset $X$. Therefore, any fast algorithm for $k$-means clustering automatically speeds up any algorithmic pipeline that uses coresets for clustering --- looking forward, we will show how our algorithm can significantly speed up coreset constructions without sacrificing approximation.

Beyond those mentioned above, many works seek to speed up $k$-means++ or Lloyd iterations by maintaining some nearest neighbor search data structures~\cite{10.1145/312129.312248, 10.5555/2073946.2073993, 10.1145/336154.336189, 1017616, 10.5555/3041838.3041857, 4270197, hamerly2010making,philbin2010scalable, 6248034, drake2013faster, pmlr-v37-ding15, pmlr-v48-bottesch16, pmlr-v48-newling16, curtin2017dual, capo2018efficient}, or by running some first-order methods~\cite{10.1145/1772690.1772862}. These techniques do not give provable guarantees on the quality of the $k$-means clustering or on the running time of their algorithms.

\paragraph{Theoretical Results.}
We give a simple randomized clustering algorithm with provable guarantees on its running time and approximation ratio without making any assumptions about the data. It has the benefit of being fast (like the algorithms in Category C above) while achieving a multiplicative error guarantee without additional additive error (like the algorithms in Category B above).
\begin{itemize}
    \item The algorithm runs in time $O(nd + n \log n)$ irrespective of $k$. It passes over the dataset once to perform data reduction, which gives the $nd$ factor plus an additive $O(n\log n)$ term to solve $k$-means on the reduced data, producing $k$ centers and cluster assignments. On sparse input datasets, the $nd$ term becomes $\nnz(X)$, where $\nnz(X)$ is the number of non-zero entries in the dataset. Thus, our algorithm runs in $O(\nnz(X) + n \log n)$ time on sparse matrices.
    
    \item The algorithm is as simple as the $k$-means++ algorithm while significantly more efficient. The approximation ratio we prove is $\poly(k)$, which is worse than the $O(\log k)$-approximation achieved by $k$-means++ but multiplicative (see the remark below on improving this to $O(\log k)$). It does not incur the additional additive errors from the fast algorithms in \cite{Bachem_Lucic_Hassani_Krause_2016,NIPS2016_d67d8ab4,10.1145/3219819.3219973}. 
\end{itemize}
Our algorithm projects the input points to a random one-dimensional space and runs an efficient $k$-means++ seeding after the projection. For the approximation guarantee, we analyze how the approximation ratio achieved after the projection can be transferred to the original points (\Cref{lm:apx-ratio}).
We bound the running time of our algorithm by efficiently implementing the $k$-means++ seeding in one dimension and analyzing the running time via a potential function argument (\Cref{lm:runtime}).
Our algorithm applies beyond $k$-means to other clustering objectives that sum up the $z$-th power of the distances for general $z\ge 1$, and our guarantees on its running time and approximation ratio extend smoothly to these settings.

\paragraph{Improving the Approximation from $\poly(k)$ to $O(\log k)$.}\label{par:improve} The approximation ratio of $\poly(k)$ may seem significantly worse than the $O(\log k)$ approximations achievable with $k$-means++. However, we can improve this to $O(\log k)$ with an additional, additive $O(\poly(kd) \cdot \log n)$ term in the running time. 
Using previous results discussed in \Cref{sec:sens} (specifically \Cref{thm:sens}),
a multiplicative $\poly(k)$-approximation suffices to construct a coreset of size $\poly(kd)$ and run $k$-means++ on the coreset. 
Constructing the coreset is simple and takes time $\poly(kd) \cdot \log n$ (by sampling from an appropriate distribution); running $k$-means++ on the coreset takes $\poly(kd)$ time (with no dependence on $n$). Combining our algorithm with coresets, we get a $O(\log k)$-approximation in $O(\nnz(X)) + O(n\log n) + \poly(kd) \cdot \log n$ time. Notably, these guarantees cannot be achieved with the additive approximations of~\cite{Bachem_Lucic_Hassani_Krause_2016,NIPS2016_d67d8ab4,10.1145/3219819.3219973}. 

\paragraph{Experimental Results.} We implemented our algorithm, as well as the lightweight coreset of~\cite{10.1145/3219819.3219973} and $k$-means++ with sensitivity sampling~\cite{DBLP:journals/corr/BravermanFL16}. We ran two types of experiments, highlighting various aspects of our algorithm. 
Our code is published on GitHub\footnote{\label{fn:github}\href{https://github.com/boredoms/prone}{\name GitHub repository: \texttt{https://github.com/boredoms/prone}}}. The two types of experiments are:

\begin{itemize}
    \item \textbf{Coreset Construction Comparison}: First, we evaluate the performance of our clustering algorithm when we use it to construct coresets. We compare the performance of our algorithm to $k$-means++ with sensitivity sampling~\cite{bachem2017practical} and lightweight coresets~\cite{10.1145/3219819.3219973}. 
    In real-world, high-dimensional data, the cost of the resulting clusters from the three algorithms is roughly the same. However, ours and the lightweight coresets can be significantly faster (ours is up to \textbf{190x} faster than $k$-means++, see Figure~\ref{fig:coreset_qualy} and Table~\ref{tab:runtime}). The lightweight coresets can be faster than our algorithm (between 3-5x); however, our algorithm is ``robust'' (achieving multiplicative approximation guarantees).\footnote{Recall that the lightweight coresets incur an additional additive error which can be arbitrarily large.} Additionally, we show that the clustering from lightweight coresets can have an arbitrarily high cost for a synthetic dataset. On the other hand, our algorithm achieves provable (multiplicative) approximation guarantees irrespective of the dataset (this is demonstrated in the right-most column of Figure~\ref{fig:coreset_qualy}).

    \item \textbf{Direct k-means++ comparison}: Second, we compare the speed and cost of our algorithm to k-means++\cite{MR2485254} as a stand-alone clustering algorithm (we also compare two other natural variants of our algorithm). Our algorithm can be up to \textbf{800x} faster than $k$-means++ for $k=5000$ and our slowest variant up to \textbf{100x} faster (Table~\ref{tab:runtime}). The cost of the cluster assignments can be significantly worse than that of $k$-means++ (see Figure~\ref{fig:cluster-cost}). 
    Such a result is expected since our theoretical results show a $\poly(k)$-approximation. The other (similarly) fast algorithms (based on approximating the $D^2$-distribution) which run in time $O(nd)$~\cite{Bachem_Lucic_Hassani_Krause_2016,NIPS2016_d67d8ab4} do not produce the cluster assignments (they only output $k$ centers). These algorithms would take $O(ndk)$ time to find the cluster assignments --- this is precisely the computational cost our algorithm avoids.
\end{itemize}
We do not compare our algorithm with~\cite{NEURIPS2020_babcff88} nor implement approximate nearest neighbor search to speed up $k$-means++ for the following reasons. The algorithm in~\cite{NEURIPS2020_babcff88} is significantly more complicated, and there is no publicly available implementation. In addition, both~\cite{NEURIPS2020_babcff88} and approximate nearest neighbor search incur additional poly-logarithmic (or even $n^{o(1)}$-factors for nearest neighbor search over $\ell_2$~\cite{AIR18}) which add significant layers of complexity to the implementation and make a thorough evaluation of the algorithm significantly more complicated. Instead, our current implementation demonstrates that a simple, one-dimensional projection and $k$-means++ on the line enables dramatic speedups to coreset constructions without sacrificing approximation quality.

\paragraph{Related Work.} Efficient algorithms for clustering problems with provable approximation guarantees have been studied extensively, with a few approaches in the literature. There are polynomial-time (constant) approximation algorithms (an exponential dependence on $k$ is not allowed) (see \cite{10.1145/2488608.2488723, BPRST15, ANSW17,GORSV22} for some of the most recent and strongest results), nearly linear time $(1\pm\eps)$-approximations with running time exponential in $k$ which proceed via coresets~(see~\cite{10.1145/1007352.1007400,doi:10.1137/070699007, FL11,FSS20, DBLP:journals/corr/BravermanFL16, bachem2017practical,CSS21, CLSS22} and references therein, as well as the surveys~\cite{AHV05, F20}), and nearly-linear time $(1\pm\eps)$-approximations in fixed / low-dimensional spaces~\cite{ARR98, KR99, T04, FRS16, CKM16, C18, 10.1145/3477541}. 
Our $O(n \log n)$-expected-time implementation of $k$-means++ seeding achieves an $O(\log k)$ expected approximation ratio for $k$-median and $k$-means in one dimension. We are unaware of previous work on clustering algorithms running in time $O(n\log n)$. 

Another line of research has been on dimensionality reduction techniques for $k$-means clustering. Dimensionality reduction can be achieved via PCA based methods~\cite{DFKVV04, FSS20, CEMMP15, SW18}, or random projection~\cite{CEMMP15,BBCGS19,MMR19}.
For random projection methods, it has been shown that the $k$-means objective is preserved up to small multiplicative factors when projecting onto $O_\eps(\log(k))$ dimensional space. Additional work has shown that dimensionality reduction can be performed in $O(\nnz(A))$ time~\cite{LST17}.
To the best of our knowledge, we are the first to show that clustering objectives such as $k$-median and $k$-means are preserved up to a $\poly(k)$ factor by one-dimensional projections.

Some works show that the $O(\log k)$ expected approximation ratio for $k$-means++ can be improved by adding local search steps after the seeding procedure \cite{pmlr-v97-lattanzi19a,pmlr-v119-choo20a}. In particular, Choo et al.\ \cite{pmlr-v119-choo20a} showed that adding $\varepsilon k$ local search steps achieves an $O(1/\varepsilon^3)$ approximation ratio with high probability.

Several other algorithmic approaches exist for fast clustering of points in metric spaces. These include density-based methods like DBSCAN~\cite{DBSCAN} and DBSCAN++~\cite{DBSCAN++} and the line of heuristics based on the Partitioning Around Medoids (PAM) approach, such as FastPAM~\cite{FastPAM}, Clarans~\cite{CLARANS}, and BanditPAM~\cite{BanditPAM}. While these algorithms can produce high-quality clustering, their running time is at least linear in the number of clusters (DBSCAN++ and BanditPAM) or superlinear in the number of points (DBSCAN, FastPAM, Clarans).

\section{Overview of Our Algorithm and Proof Techniques}
\label{sec:overview}
Our algorithm, which we call \name (PRojected ONE-dimensional clustering), takes a random projection onto a one-dimensional space, sorts the projected (scalar) numbers, and runs the $k$-means++ seeding strategy on the projected numbers. By virtue of its simplicity, the algorithm is scalable and effective at clustering massive datasets. 
More formally, \name receives as input a dataset of $n$ points in $\R^d$, a parameter $k \in \N$ (the number of desired clusters), and proceeds as follows:

\begin{enumerate}
\item\label{en:step-project} Sample a random vector $v\in \R^d$ from the standard Gaussian distribution and project the data points to one dimension along the direction of $v$. That is, we compute $x_i' = \langle x_i,v\rangle\in \R$ in time $O(\nnz(X))$ by making a single pass over the data, effectively reducing our dataset to the collection of one-dimensional points $x_1',\ldots,x_n' \in \R$.
\item\label{en:step-one-dim} Run $k$-means++ seeding on $\smash{x_1',\ldots,x'_n}$ to obtain $k$ indices $j_1,\ldots,j_k\in [n]$ indicating the chosen centers $\smash{x_{j_1}',\ldots,x_{j_k}'}$ and an assignment $\sigma:[n]\to [k]$ assigning point $x_i'$ to center $\smash{x_{j_{\sigma(i)}}'}$. Even though $k$-means++ seeding generally takes $O(nk)$ time in one dimension, we give an efficient implementation, leveraging the fact that points are one-dimensional, which runs in $O(n \log n)$ expected time, independent of $k$. 
A detailed algorithm description is in \cref{sec:runtime}.
\item The one-dimensional $k$-means++ algorithm produces a collection of $k$ centers $x_{j_1},\ldots,x_{j_k}$, as well as the assignment $\sigma$ mapping each point $x_i$ to the center $x_{j_{\sigma(i)}}$. For each $\ell \in [k]$, 
we update the cluster center for cluster $\ell$ to be the center of mass of all points assigned to $x_{j_{\ell}}$.
\end{enumerate}
While the algorithm is straightforward, the main technical difficulty lies in the analysis. In particular, our analysis (1) bounds the approximation loss incurred from the one-dimensional projection in Step~\ref{en:step-project} 
and (2) shows that we can implement Step~\ref{en:step-one-dim} in $O(n \log n)$ expected time, as opposed to $O(nk)$ time. 
We summarize the theoretical contributions in the following theorems.

\begin{theorem}
\label{thm:runtime}
The algorithm \name has expected running time $O(\nnz(X) + n\log n)$ on any dataset $\smash{X = \{ x_1,\ldots,x_n\} \subset \R^d}$. Moreover, for any $\delta\in (0,1/2)$ and any dataset $X$, with probability at least $1-\delta$, the algorithm runs in time $O(\nnz(X) + n\log (n/\delta))$.
\end{theorem}
\begin{theorem}
\label{thm:apx-ratio}
The algorithm \name achieves an $\widetilde{O}(k^4)$ approximation ratio for the $k$-means objective with probability at least $0.9$.
\end{theorem}
To our knowledge, \name is the first algorithm for $k$-means running in time $O(nd + n\log n)$ for arbitrary $k$. As mentioned in \hyperref[par:improve]{the paragraph on improving the competitive ratio}, we obtain the following corollary of \Cref{thm:runtime,thm:apx-ratio} using a two-stage approach with a coreset:
\begin{corollary}
\label{thm:improved-apx}
By using \name as the $\alpha$-approximation algorithm in Theorem~\ref{thm:sens} and running $k$-means++ on the resulting coreset, we obtain an algorithm with an approximation ratio of $O(\log k)$ that runs in time $O(\nnz(X) + n\log n + \poly(kd) \log n)$, with constant success probability.
\end{corollary}

The proofs of \Cref{thm:runtime,thm:apx-ratio} can be found in \Cref{sec:apx-ratio,sec:runtime}, where we also generalize them beyond $k$-means to clustering objectives that sum up the $z$-th power of Euclidean distances for general $z\ge 1$. The following subsections give a high-level overview of the main techniques we develop to prove our main theorems above.

\subsection{Efficient Seeding in One Dimension}

The $k$-means++ seeding procedure has $k$ iterations, where a new center is sampled in each iteration. Since a new center may need to update $\Omega(n)$ distances to maintain the $D^2$ distribution, which samples each point with probability proportional to its distance to its closest center, a naive analysis leads to a running time of $O(nk)$. A key ingredient in the proof of \Cref{thm:runtime} is showing that, for one-dimensional datasets, $k$-means++ only needs to make $O(n\log n)$ updates, irrespective of $k$. 
\begin{lemma}
\label{lm:runtime}
The $k$-means++ seeding procedure can be implemented in expected time $O(n\log n)$ in one dimension.
Moreover, for any $\delta\in (0,1/2)$, with probability at least $1-\delta$, the implementation runs in time $O(n\log (n/\delta))$.
\end{lemma}

The intuition of the proof is as follows: Since points are one-dimensional, we always maintain them in sorted order. In addition, each data point $x_i$ will maintain its center assignment and distance $p_i$ to the closest center. By building a binary tree over the sorted points (where internal nodes maintain sums of $p_i^2$'s), it is easy to sample a new center from the $D^2$ distribution in $O(\log n)$ time. The difficulty is that adding a new center may result in changes to $p_i$'s of multiple points $x_i$, so the challenge is to bound the number of times these values are updated (see Figure~\ref{fig:seeding1d} below).

\begin{figure}[h]
\centering
\includegraphics[width = 0.8\textwidth]{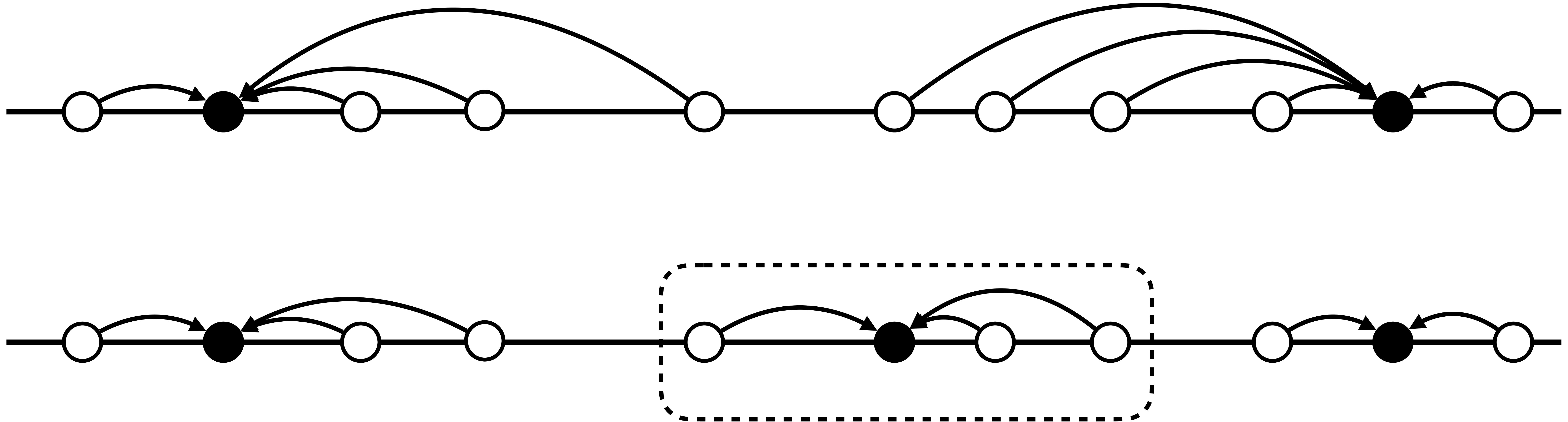}
\caption{From the top to the bottom, a new center (black circle) is chosen. Every point has an arrow pointing to its closest center. The points in the dashed box are the ones that require updates.}
\label{fig:seeding1d}
\end{figure}

To bound the total running time, we leverage the one-dimensional structure. Observe that, for a new center, the updated points lie in a contiguous interval around the newly chosen center.
Once a center is chosen, the algorithm scans the points (to the left and the right) until we reach a point that does not need to be updated.
This point identifies that points to the other side of it need not be updated, so we can get away
without necessarily checking all $n$ points (see \Cref{fig:seeding1d}).
Somewhat surprisingly, when sampling centers from the $D^2$-distribution, 
the expected number of times that each point will be updated is only $O(\log n)$, which implies a bound of $O(n\log n)$ on the total number of updates in expectation. The analysis of the fact that each point is updated $O(\log n)$ times is non-trivial and uses a carefully designed potential function (Lemma~\ref{lm:eta}). 

\subsection{Approximation Guarantees from One-Dimensional Projections}

Our proof of \Cref{thm:apx-ratio} builds on a line of work studying randomized dimension reduction for clustering problems \cite{boutsidis2014randomized,CEMMP15,BBCGS19,MMR19}. Prior work studied randomized dimension reduction for accurate $(1\pm \epsilon)$-approximations. Our perspective is slightly different; 
we restrict ourselves to one-dimensional projections and give an upper bound on the distortion. 

For any dataset $x_1, \dots, x_n \in \R^d$, a projection to a random lower-dimensional space affects the pairwise distance between the projected points in a predictable manner --- the Johnson-Lindenstrauss lemma which projects to $O(\log n)$ dimensions being a prime example of this fact. 
When projecting to just one dimension, however, pairwise distances will be significantly affected (by up to $\poly(n)$-factors). Thus, a naive analysis will give a $\poly(n)$-approximation for $k$-means.
To improve a $c$-approximation to a $O(\log k)$-approximation, one needs a coreset of size roughly $\poly(c / \log k)$. This bound becomes vacuous when $c$ is polynomial in $n$ since there are at most $n$ dataset points.

However, although many pairwise distances are significantly distorted, we show that the $k$-means cost is only affected by a $\poly(k)$-factor. At a high level, this occurs because the $k$-means cost optimizes a sum of pairwise distances (according to a chosen clustering). The individual summands, given by pairwise distances, will change significantly, but the overall sum does not. Our proof follows the approach of~\cite{CEMMP15}, which showed that (roughly speaking) pairwise distortion of the $k$ optimal centers suffices to argue about the $k$-means cost. The $k$ optimal centers will incur maximal pairwise distortion $\poly(k)$ when projected to one dimension (because there are only $O(k^2)$ pairwise distances among the $k$ centers).
This allows us to lift an $r$-approximate solution after the projection to an $O(k^4r)$-approximate solution for the original points.
\begin{lemma}[Informal]
\label{lm:apx-ratio}
For any set $X$ of points in $\R^d$, the following occurs with probability at least $0.9$ over the choice of a standard Gaussian vector $v\in\R^d$. Letting $X' \subset \R$ be the one-dimensional projection of $X$ onto $v$, any $r$-approximate $k$-means clustering of $X'$ gives an $O(k^4 r)$-approximate clustering of $X$ with the same clustering partition. 
\end{lemma}

\section{Preliminaries}\label{sec:prelims}

In this work, we always consider datasets $X = \{ x_1,\dots, x_n \} \subset \R^d$ of high-dimensional vectors, and we will measure their distance using the Euclidean ($\ell_2$) distance. Below, we define $(k,z)$-clustering. This problem introduces a parameter $z \geq 1$, which measures the \emph{sensivity to outliers} (as $z$ grows, the clusterings become more sensitive to points furthest from the cluster center). The case of $k$-means corresponds to $z = 2$, but other values of $z$ capture other well-known clustering objectives, like $k$-median (the case of $z=1$).

\begin{definition}[$(k,z)$-Clustering] \label{def:k-z-cluster}
Consider a dataset $X = \{ x_1,\dots, x_n \} \subset \R^d$, a desired number of clusters $k \in \N$, and a parameter $z \geq 1$. For a set of $k$ centers $C = \{ c_1, \dots, c_k \} \subset \R^d$, let $\cost_z(X, C)$ denote the cost of using the center set $C$ to cluster $X$, i.e.,
\[ \cost_z(X, C) = \sum_{i =1}^n \min_{j \in [k]} \| x_i - c_j \|_2^z. \]
We let $\opt_{k,z}(X)$ denote the optimal cost over all choices of $C = \{c_1,\ldots,c_k\} \subset \R^d$:
\[
\opt_{k,z}(X) = \inf_{\substack{C \subset \R^d \\ |C| \leq k}}\cost_z(X,C).
\]
\end{definition}

A $(k,z)$-clustering algorithm has the following specifications.
The algorithm receives as input a dataset $X = \{ x_1, \dots, x_n \} \subset \R^d$, as well as two parameters $k \in \N$ and $z \geq 1$. After it executes, the algorithm should output a set of $k$ centers $C = \{ c_1,\dots, c_k \} \subset \R^d$ as well as an assignment $\sigma:[n]\to [k]$ mapping each point $x_i$ to a center $c_{\sigma(i)}$.

We measure the quality of the solution $(C,\sigma)$ using the ratio between its $(k,z)$-clustering cost $\cost_z(X, C, \sigma)$ and the optimal cost $\opt_{k,z}(X)$, where
\begin{equation}
\label{eq:assignment-cost}
\cost_z(X, C, \sigma) := \sum_{i=1}^n \|x_i - c_{\sigma(x_i)}\|_2^z.
\end{equation}
For any $\sfD > 1$, an algorithm that produces a $\sfD$-approximation to $(k,z)$-clustering should guarantee that $\cost_z(X, C,\sigma)$ is at most $\sfD \cdot \opt_{k,z}(X)$. For a randomized algorithm, the guarantee should hold with large probability (referred to as the success probability) for any input dataset.

\subsection{\texorpdfstring{$k$-Means++ Seeding}{k-Means++ Seeding}}

The $k$-means++ seeding algorithm is a well-studied algorithm introduced in \cite{MR2485254}, and it is an important component of our algorithm. 
Below, we describe it for general $z \ge 1$, not necessarily $z = 2$.

\begin{definition}[$k$-means++ seeding, for arbitrary $z \geq 1$]
\label{def:k-means++}
Given $n$ data points $x_1,\ldots,x_n\in \R^d$, the $k$-means++ seeding algorithm produces a set of $k$ centers, $x_{\ell_1},\dots, x_{\ell_k}$ with the following procedure:
\begin{enumerate}
\item Choose $\ell_1$ uniformly at random from $[n]$.
\item\label{en:step-2} For $t = 2,\ldots,k$, sample $\ell_t$ as follows. For every $i \in [n]$, let $p_i$ denote the Euclidean distance from $x_i$ to its closest point among $x_{\ell_1},\ldots,x_{\ell_{t - 1}}$. Sample $\ell_t$ from $[n]$ so that the probability $\Pr[\ell_t = i]$ is proportional to $p_i^z$ for every $i\in [n]$. That is,
\[
\Pr[\ell_t = i] = \frac{p_i^z}{\sum_{i'\in [n]}p_i^z}.
\]
In the context of $k$-means (i.e., when $z =2$), the distribution is known as the $D^2$ distribution.
\item Output $x_{\ell_1},\ldots,x_{\ell_k}$.
\end{enumerate}
\end{definition}

In the above description, Step~\ref{en:step-2} of $k$-means++ needs to maintain, for each dataset point $x_i$, the Euclidean distance $p_i$ to its closest center among the centers selected before the current iteration. This step is implemented by making an entire pass over the dataset for each of the $k-1$ iterations of Step~\ref{en:step-2}, leading to an $O(ndk)$ running time.

\begin{theorem}[\cite{MR2485254}, Theorem 3 in \cite{Waingarten}]\label{thm:k-means++}
For $x_1,\ldots,x_n\in \R^d$, let $X = \{x_1,\ldots,x_n\}$ be the input to the $k$-means++ seeding algorithm.
For the output $x_{\ell_1},\ldots,x_{\ell_k}$ of the $k$-means++ seeding algorithm, define $C = \{x_{\ell_1},\ldots,x_{\ell_k}\}$. Then
\[
\E[\cost_z(X,C)] = O(2^{2z}\log k)\cdot \opt_{k,z}(X).
\]
\end{theorem}

\subsection{Coresets via Sensitivity Sampling}
\label{sec:sens}
One of our algorithm's applications is constructing coresets for $(k,z)$-clustering. We give a formal definition and describe the primary technique for building coresets.

\begin{definition}
Given a dataset $X = \{ x_1,\dots, x_n \} \subset \R^d$, as well as parameters $k \in \N$, $z \geq 1$ and $\eps > 0$, a (strong) $\eps$-coreset for $(k,z)$-clustering is specified by a set of points $Y \subset \R^d$ and a weight function $w \colon Y \to \R_{\geq 0}$, such that, for every set $C = \{ c_1,\dots, c_k \} \subset \R^d$,
\begin{align*}
(1-\eps) \cdot \cost_z(X, C) \leq \sum_{y\in Y} w(y) \cdot \min_{j \in [k]} \| y - c_j\|_2^z \leq (1+\eps) \cdot \cost_z(X, C).
\end{align*}
\end{definition}

Coresets are constructed via ``sensitivity sampling,'' a technique that, given an approximate clustering of a dataset $X$, produces a probability distribution such that sampling enough points from this distribution results in a coreset. 

\begin{definition}[Sensitivity Sampling]\label{def:sens}
Consider a dataset $X = \{ x_1,\dots, x_n \} \subset \R^d$, as well as parameters $k \in \N$, $z \geq 1$. For a centet set $C = \{ c_1,\dots, c_k \} \subset \R^d$ and assignment $\sigma \colon [n] \to [k]$, let $X_{j} = \{ x_i : \sigma(i) = j \}$. We let $\calD$ be a distribution supported on $X$ where
\begin{align*}
    \Prx_{\bx \sim \calD}\left[\bx = x_i \right] \propto \dfrac{\|x_i - c_{\sigma(i)}\|_2^z}{\sum_{j=1}^n \|x_j - c_{\sigma(j)}\|_2^z} + \frac{1}{|X_{\sigma(i)}|}.
\end{align*}
\end{definition}

The main theorem that we will use is given below, which shows that given a center set and an assignment that gives an $\alpha$-approximation to $(k,z)$-clustering, one may sample from the distribution $\calD$ defined about in order to generate a coreset with high probability.

\begin{theorem}[\cite{DBLP:journals/corr/BravermanFL16}]\label{thm:sens}
For any dataset $X = \{ x_1,\dots, x_n \} \subset \R^d$ and any parameters $k \in \N$ and $z \geq 1$, suppose that $C = \{ c_1,\dots ,c_k\} \subset \R^d$ and $\sigma \colon [n] \to [k]$ is a $\alpha$-approximation to $(k,z)$-clustering, i.e.,
\[ \sum_{i=1}^n \|x_i - c_{\sigma(i)}\|_2^z \leq \alpha\, \opt_{k,z}(X). \]
Letting $\calD$ denote the distribution specified in Definition~\ref{def:sens}, the following occurs with high probability. 
\begin{itemize}
    \item We let $\by_1,\dots, \by_s$ denote independent samples from $\calD$, and $w(\by_i)$ be the inverse of the probability that $\by_i$ is sampled according to $\calD$. We set $s \geq \poly(kd \cdot \alpha \cdot 2^z/\eps)$.
    \item The set $\bY = \{ y_1,\dots, y_s \}$ with weights $w$ is an $\eps$-coreset for $(k,z)$-clustering.
\end{itemize}
\end{theorem}

\subsection{A Simple Lemma}

We will repeatedly use the following simple lemma.
\begin{lemma}\label{lem:jensen}
Let $a, b \in \R_{\geq 0}$ be any two numbers and $z \geq 1$. Then,
$(a + b)^z \leq 2^{z-1} a^z + 2^{z-1} b^z$.
\end{lemma}
\begin{proof}
The function $\phi(t) = t^z$ is convex for $z \geq 1$, so Jensen's inequality implies $\phi((a + b)/2) \leq (1/2) \phi(a) + (1/2) \phi(b)$.
\end{proof}

\section{Approximation Guarantees from One-Dimensional Projections}
\label{sec:apx-ratio}
In this section, we prove \Cref{thm:apx-ratio} (rather, the generalization of \Cref{thm:apx-ratio} to any $z \geq 1$) by analyzing the random one-dimensional projection step in our algorithm. In order to introduce some notation, let $X = \{ x_1,\dots, x_n \} \subset \R^d$ be a set of points, and for a partition of $X$ into $k$ sets, $(Y_1,\dots, Y_k)$, we let the
$(k,z)$-\emph{clustering cost of $X$ with the partition $(Y_1,\dots, Y_k)$} be
\begin{align}
\cost_z(Y_1,\dots, Y_k) = \sum_{i=1}^k \min_{c \in \R^d} \sum_{x \in Y_i} \| x - c\|_2^z \label{eq:cluster-cost}
\end{align}
and call the $k$ points $c$ selected as minima a set of centers realizing the $(k,z)$-clustering cost of $(Y_1,\dots, Y_k)$. We note that (\ref{eq:cluster-cost}) is a cost function for $(k,z)$-clustering, but it is different from Definition~\ref{def:k-z-cluster}. In Definition~\ref{def:k-z-cluster}, the emphasis is on the set of $k$ centers $C = \{ c_1,\dots, c_k\}$, and the induced set of clustering of $X$, i.e., the partition $(Y_1,\dots, Y_k)$ given by assigning points to the closest center, is only implicitly specified by the set of centers. On the other hand, (\ref{eq:cluster-cost}) emphasizes the clustering $(Y_1,\dots, Y_k)$, and the set of $k$ centers implicitly specified by $(Y_1,\dots, Y_k)$. The optimal set of centers and the optimal clustering will achieve the same cost; however, our proof will mostly consider the clustering $(Y_1,\dots, Y_k)$ as the object to optimize. Shortly, we will sample a (random) dimensionality reduction map $\bPi \colon \R^d \to \R^t$ and seek bounds for $t = 1$. We will write $\cost_z(\bPi(Y_1),\dots, \bPi(Y_k))$ for the cost of clustering the points after applying the dimensionality reduction map $\bPi$ to the partition $Y_1,\dots, Y_k$. Namely, we write
\[ \cost_z(\bPi(Y_1),\dots, \bPi(Y_k)) = \sum_{i=1}^k \min_{c \in \R^t} \sum_{x \in Y_i} \| \bPi(x) - c \|_2^z. \]

\begin{definition}
For a set of points $X = \{ x_1,\dots, x_n \} \subset \R^d$, we use $X_1^*,\dots, X_k^*$ of $X$ to denote the partition of $X$ with minimum $(k,z)$-clustering cost and we use $C^* = \{ c_1^*,\dots, c_k^* \} \subset \R^d$ to denote a set of $k$ centers which realizes the $(k,z)$-clustering cost of $X_1^*,\dots, X_k^*$, i.e., the set of centers which satisfies
\begin{align*}
\cost_z(X_1^*, \dots, X_k^*) &= \sum_{i=1}^k \sum_{x \in X_{i}^*} \| x - c_i^*\|_2^z. 
\end{align*}
By slight abuse of notation, we also let $c^* \colon X \to C^*$ be the map which sends every point of $X$ to its corresponding center (i.e., if $x \in X_i^*$, then $c^*(x)$ is the point $c_i^*$).
\end{definition}

We prove the following lemma, which generalizes Lemma~\ref{lm:apx-ratio} from $k$-means to $(k,z)$-clustering (recall that $k$-means corresponds to the case of $z = 2$).
\begin{lemma}[Effect of One-Dimensional Projection on $(k,z)$-Clustering]\label{lm:apx-formal}
For $n, d, k \in \N$ and $z \geq 1$, let $X = \{ x_1,\dots, x_n \} \subset \R^d$ be an arbitrary dataset. We consider the (random) linear map $\bPi \colon \R^d \to \R$ given by sampling $\bg \sim \calN(0, I_d)$ and setting
\[ \bPi(x) = \langle x, \bg \rangle. \]
With probability at least $0.9$ over $\bg$, the following occurs:
\begin{itemize}
    \item We consider the projected dataset $\bX' = \{ \bx_1', \dots, \bx_n' \} \subset \R$ be given by $\bx_i' = \bPi(x_i)$, and
    \item For any $r\geq1$, we let $(Y_1,\dots Y_k)$ denote any partition of $X$ satisfying
    \[ \cost_{z}(\bPi(Y_1),\dots, \bPi(Y_k)) \leq r \cdot \min_{c_1,\dots, c_k \in \R} \sum_{i=1}^n \min_{j \in [k]} |\bx_i' - c_j|^z.\]
\end{itemize}
Then,
\[ \cost_z(Y_1,\dots, Y_k) \leq 2^{O(z)} \cdot k^{2z} \cdot r \cdot \opt_{k,z}(X). \]
\end{lemma}

By setting $z = 2$, we obtain the desired bound from Lemma~\ref{lm:apx-ratio}. We can immediately see that, from  Lemma~\ref{lm:apx-formal}, and the approximation guarantees of $k$-means++ (or rather, its generalization to $z \geq 1$) in Theorem~\ref{thm:k-means++}, we obtain our desired approximation guarantees. Below, we state the generalization of Theorem~\ref{thm:apx-ratio} to all $z \geq 1$ and, assuming Lemma~\ref{lm:apx-formal}, its proof.

\begin{theorem}[Generalization of Theorem~\ref{thm:apx-ratio} to $z \geq 1$]\label{thm:apx-gen-z}
    For $n, d, k \in \N$ and $z \geq 1$, let $X = \{ x_1,\dots, x_n \} \subset \R^d$ be an arbitrary dataset. We consider the following generalization of our algorithm \name:
    \begin{enumerate}
        \item Sample a random Gaussian vector $\bg \sim \calN(0, I_d)$ and consider the projection $\bX' = \{\bx_1',\dots,\bx_n'\}$ given by $\bx_i' = \bPi(x_i)$, for $\bPi(x) = \langle x, \bg\rangle \in \R$.
        \item Execute the (generalization of the) $k$-means++ seeding strategy for $z \geq 1$ of Definition~\ref{def:k-means++} with the dataset $\bX' \subset \R$, and let $\bx_{j_1}',\dots, \bx_{j_k}' \in \R$ denote the centers and $(\bY_1,\dots, \bY_k)$ denote the partition of $X$ specifying the $k$ clusters found.
        \item Output the clustering $(\bY_1,\dots, \bY_k)$, and the set of centers $\bc_1,\dots, \bc_k \in \R^d$ where
        \[ \bc_{\ell} = \mathop{\Ex}_{\bx \sim \bY_{\ell}}\left[ \bx\right] \in \R^d.\]
    \end{enumerate}
    Then, with probability at least $0.8$ over the execution of the algorithm, 
    \begin{align*}
        \cost_z(\bY_1,\dots, \bY_k) \leq \sum_{\ell=1}^k \sum_{x \in Y_{\ell}} \| x - \bc_{\ell} \|_2^z \leq 2^{O(z)} \cdot k^{2z} \cdot \log k \cdot \opt_{k,z}(X).
    \end{align*}
\end{theorem}

\begin{proof}[Proof of Theorem~\ref{thm:apx-gen-z} assuming Lemma~\ref{lm:apx-formal}]
    We consider the case (over the randomness in the execution of the algorithm) that:
    \begin{enumerate}
        \item\label{en:event-1} The conclusions of Lemma~\ref{lm:apx-formal} hold for the projected dataset $\bX'$ (which happens with probability at least $0.9$) by Lemma~\ref{lm:apx-formal}.
        \item\label{en:event-2} The execution of the generalization $k$-means++ seeding strategy on $\bX'$ (from Definition~\ref{def:k-means++}) produces a set of centers $\{\bx_{j_1}', \dots, \bx_{j_k}'\}\subset \R$ which cluster $\bX'$ with cost at most $O(2^{2z} \log k) \cdot \opt_{k,z}(\bX')$ (which also happens with probability $0.9$ by Markov's inequality).
    \end{enumerate}
    By a union bound, both hold with probability at least $0.8$. We now use Lemma~\ref{lm:apx-formal} to upper bound the cost of the clustering $(\bY_1,\dots, \bY_k)$. The first inequality is trivial; suppose we let $\hat{c}_{1}^*,\dots, \hat{c}_{\ell}^* \in \R^d$ be the centers which minimize for each $\ell \in [k]$
    \[ \min_{\hat{c}_{\ell} \in \R^d} \sum_{x \in \bY_{\ell}} \| x - \hat{c}_{\ell} \|_2^z = \sum_{x\in \bY_{\ell}} \| x - \hat{c}_{\ell}^*\|_2^z. \]
    Then, we trivially have
    \[ \cost_z(\bY_1,\dots, \bY_k) = \sum_{\ell=1}^k \sum_{x \in \bY_{\ell}} \| x - \hat{c}_{\ell}^* \|_2^z \leq \sum_{\ell=1}^k \sum_{x \in \bY_{\ell}} \| x - \bc_{\ell} \|_2^z.\]
    Furthermore, we can also show a corresponding upper bound. For each $\ell \in [k]$, recall that $\bc_{\ell} \in \R^d$ is the center of mass of $\bY_k$, so we can apply the triangle inequality and Lemma~\ref{lem:jensen}
    \begin{align*}
    \sum_{x \in \bY_{\ell}} \| x - \bc_{\ell}\|_2^z &\leq 2^{z-1} \sum_{x \in \bY_{\ell}} \| x - \hat{c}_{\ell}^*\|_2^z + 2^{z-1} |\bY_{\ell}| \cdot \|\hat{c}_{\ell}^* - \mathop{\Ex}_{\bx \sim \bY_{\ell}}[\bx]\|_2^z \\
                &\leq 2^{z-1} \sum_{x \in \bY_{\ell}} \| x - \hat{c}_{\ell}^*\|_2^z + 2^{z-1} |\bY_{\ell}| \cdot \mathop{\Ex}_{\bx \sim \bY_{\ell}}\left[\|x - \hat{c}_{\ell}^*\|_2^z \right],
    \end{align*}
    where the second inequality is Jensen's inequality, since $\phi(x) = \|\hat{c}_{\ell}^* - x\|_2^z$ is convex for $z \geq 1$. Thus, we have upper-bounded
    \[ \sum_{x \in \bY_{\ell}} \|x - \bc_{\ell} \|_2^z \leq 2^{z} \sum_{x \in \bY_{\ell}} \| x - \hat{c}_{\ell}^* \|_2^z,\]
    and therefore
    \begin{align} 
    \sum_{\ell=1}^k \sum_{x \in \bY_{\ell}} \| x - \bc_{\ell} \|_2^z \leq 2^{z} \cdot \cost_z(\bY_1,\dots, \bY_{k}). \label{eq:costs-relate}
    \end{align}
    The final step involves relating $\cost_z(\bY_1,\dots, \bY_k)$ using the conclusions of Lemma~\ref{lm:apx-formal}. Notice that our algorithm produces the clustering $(\bY_1,\dots, \bY_k)$ of $\bX'$ which is specified by letting
    \[ \bY_{\ell} = \left\{ x_i \in X : \forall j' \in [k], |\bx_i' - \bx_{j_{\ell}}'| \leq |\bx_i' - \bx_{j'}|^z \right\},\]
    and by the event (\ref{en:event-2}), we have $\cost_z(\bPi(\bY_1),\dots ,\bPi(\bY_{k})) \leq O(2^{2z} \log k) \cdot \opt_{k,z}(\bX')$. By event (\ref{en:event-1}), Lemma~\ref{lm:apx-formal} implies that $\cost_z(\bY_1,\dots, \bY_{k}) \leq 2^{O(z)} \cdot k^{2z} \cdot O(2^{2z} \log k) \cdot \opt_{k,z}(X)$. Combined with (\ref{eq:costs-relate}), we obtain our desired bound.
\end{proof}

\subsection{Proof of Lemma~\ref{lm:apx-formal}}

We now turn to the proof of Lemma~\ref{lm:apx-formal}, where our analysis will proceed in two steps. 
First, we assume a fixed dimensionality reduction map $\Pi \colon \R^d \to \R^t$, which satisfies two geometrical conditions on $\Pi$. Under these conditions, we show how to ``lift'' an approximate clustering of the mapped points in $\R^t$ to an approximate clustering of the original dataset in $\R^d$ at the cost of weakening the approximation ratio. Then, we show that a simple one-dimensional projection $\bPi \colon \R^d \to \R$ given by $\bPi(x) = \langle x, \bg\rangle$, for $\bg$ being sampled from a $d$-dimensional standard Gaussian, satisfies the geometrical conditions of our lemma. 

\begin{lemma}\label{lem:approx-bound}
Let $X = \{ x_1,\dots, x_n \} \subset \R^d$ and $\Pi \colon \R^d \to \R^t$ be a linear map. Let $C = \{ c_1^*,\dots, c_k^*\} \subset \R^d$ denote the set of centers minimizing $\cost_z(X, C)$, and $(X_1^*,\dots, X_k^*)$ denote the optimal $(k,z)$-clustering, and suppose that for the parameters $\sfD_1, \sfD_2, \sfD_3 \geq 1$, the following conditions hold:
\begin{itemize}
\item \emph{\textbf{Centers Don't Contract}}: Every $i, j \in [k]$ satisfies
\begin{align*}
\| c_i^* - c_j^* \|_2 \leq \sfD_1 \cdot \| \Pi(c_i^*) - \Pi(c_j^*) \|_2.
\end{align*} 
\item \emph{\textbf{Cost of $(X_1^*,\dots, X_k^*)$ does not Increase}}: We have that
\begin{align*}
 \sum_{i=1}^k \sum_{x \in X_i^*} \| \Pi(x) - \Pi(c_i^*)\|_2^z \leq \sfD_2 \cdot \cost_z(X_1^*, \dots, X_k^*).
\end{align*}
\item \emph{\textbf{Approximately Optimal $(\Pi(Y_1),\dots, \Pi(Y_k))$}}: The partition $(Y_1,\dots, Y_k)$ of $X$ is $\sfD_3$ -approximately optimal for $\Pi(X)$, i.e.,
\[ \cost_z(\Pi(Y_1), \dots, \Pi(Y_k)) \leq \sfD_3 \cdot \min_{c_1,\dots, c_k \in \R^t} \sum_{x \in X} \min_{j \in[k]} \| \Pi(x) - c_j \|_2^z.  \]
\end{itemize}
Then, 
\[ \cost_z(Y_1,\dots, Y_k) \leq \left(2^{z-1} + 2^{3z-2} \sfD_1^z \sfD_2 (1 + \sfD_3) \right) \cdot \cost_z(X_1^*,\dots, X_k^*). \]
\end{lemma}
Before starting the proof of Lemma~\ref{lem:approx-bound}, we show that projecting points onto a random Gaussian vector gives the first two desired guarantees of the above lemma with $\sfD_1 := (k^2 / \delta)$ and $\sfD_2 := 2^{O(z)}/\delta$ with probability at least $1-\delta$. The first lemma that we state below shows that the first condition of Lemma~\ref{lem:approx-bound} is satisfied with high probability, and the second lemma that the second condition of Lemma~\ref{lem:approx-bound} is satisfied with high probability.

\begin{lemma}[Centers Don't Contract]\label{lem:centers-not-contract}
Let $C = \{ c_1,\dots, c_k \} \subset \R^d$ denote any collection of $k$ points and let $\bPi \colon \R^d \to \R$ be a random map given by
\[ \bPi(x) = \langle x, \bg\rangle\]
for a randomly chosen vector $\bg \sim \calN(0, I_d)$. Then, with probability at least $1-\delta$ over $\bg$, every $i,j \in [k]$ satisfies
\begin{align*}
\left(\frac{\delta}{k^2}\right) \cdot \| c_i - c_j\|_2 \leq \| \bPi(c_i) - \bPi(c_j) \|_2.
\end{align*}
\end{lemma}

\begin{lemma}[Cost of $(X_1^*,\dots, X_k^*)$ does not Increase]\label{lem:cost-increase}
Let $X = \{ x_1,\dots, x_n \} \subset \R^d$ and let $X_1^*,\dots, X_k^*$ be the partition of $X$, and $c_1^*,\dots, c_k^* \in \R^d$ be the centers which minimize the $(k,z)$-clustering cost of $X$. Then, with probability a least $1 - \delta$, 
\begin{align*}
\sum_{i=1}^k \sum_{x \in X_i^*} \| \bPi(x) - \bPi(c_i^*) \|_2^z \leq \left(2^{O(z)} / \delta \right) \sum_{i=1}^k \sum_{x \in X_i^*} \| x - c_i^*\|_2^z.
\end{align*}
\end{lemma}

\begin{proof}[Proof of Lemma~\ref{lm:apx-formal} assuming Lemma~\ref{lem:approx-bound}, Lemma~\ref{lem:centers-not-contract} and Lemma~\ref{lem:cost-increase}]
We will apply Lemma~\ref{lem:approx-bound} by letting $\delta$ be a small enough constant (say, $\delta = 0.01$) to take a union bound. Lemma~\ref{lem:centers-not-contract} implies the first condition with $\sfD_1 = O(k^2)$ and Lemma~\ref{lem:cost-increase} implies the second condition with $\sfD_2 = 2^{O(z)}$. Finally, the second assumption of Lemma~\ref{lm:apx-formal} sets $r = \sfD_3$, from which we derive the conclusion.
\end{proof}

We now prove Lemma~\ref{lem:centers-not-contract}, Lemma~\ref{lem:cost-increase}. Lemma~\ref{lem:approx-bound} is proved in Subsection~\ref{sec:apr-bound-proof}.

\begin{proof}[Proof of Lemma~\ref{lem:centers-not-contract}]
The proof relies on the 2-stability property of the Gaussian distribution. Namely, if we let $z \in \R^d$ be an arbitrary vector and we sample a standard Gaussian vector $\bg \sim \calN(0, I_d)$, the (scalar) random variable $\langle z, \bg \rangle$ is distributed like $\|z\|_2 \cdot \bg'$, where $\bg \sim \calN(0, 1)$.
Using the $2$-stability of the Gaussian distribution for every $i, j \in [k]$, we have that $\| \bPi(c_i) - \bPi(c_j) \|_2^2$ is distributed as $(\bg')^2 \| c_i - c_j \|_2^2$, where $\bg'$ is distributed as a (one-dimensional) Gaussian $\calN(0, 1)$. Thus, by a union bound, the probability that there exists a pair $i, j \in [k]$, which satisfies $\| \bPi(c_i) - \bPi(c_j)\|_2^2 < \alpha^2 \cdot \| c_i - c_j\|_2^2$ is at most $k^2$ times the probability that a Gaussian random variable lies in $[-\alpha, \alpha]$, and this probability is easily seen to be less than $\alpha$. Setting $\alpha = \delta / k^2$ gives the desired lemma.
\end{proof}

\begin{proof}[Proof of Lemma~\ref{lem:cost-increase}]
Similarly to the proof of Lemma~\ref{lem:centers-not-contract}, we have that $\| \bPi(x) - \bPi(c_i^*)\|_2^z$ is distributed as $|\bg'|^{z} \cdot \| x - c_i^*\|_2^z$, where $\bg'$ is distributed as a (one-dimensional) Gaussian $\calN(0, 1)$. By linearity of expectation,
\begin{align*}
\Ex_{\bPi \sim \calJ_d}\left[ \sum_{i=1}^k \sum_{x \in X_i^*} \| \bPi(x) - \bPi(c_i^*)\|_2^z\right] &= \sum_{i=1}^k \sum_{x \in X_i^*} \mathop{\Ex}_{\bg' \sim \calN(0, 1)}\left[ |\bg_i'|^z \right] \cdot \| x - c_i^*\|_2^z.
\end{align*}
To conclude, note that for $z \geq 1$, there is some $\alpha > 1$ such that $\alpha z$ is an even integer and $\alpha \leq 2$. Thus, by Jensen's inequality and the fact that $f(x) = x^{1/\alpha}$ is concave we can write
\begin{align*}
\mathop{\Ex}_{\bg' \sim \calN(0, 1)}\left[ |\bg'|^z \right] \leq \left( \Ex\left[ (\bg')^{\alpha z} \right]\right)^{1/\alpha}
\end{align*}

Now note that 
all odd moments of the Gaussian distribution are zero by symmetry.
Thus, for the moment generating function $\Ex[e^{\bg'}]$ it holds that
\[ \Ex[ e^{\bg'}] = \sum_{k=0}^{\infty} \frac{1}{(2k)!} \cdot \Ex[(\bg')^{2k}].
\]

As $\Ex[e^{\bg'}] \leq e^{1/2}$ it follows that
\[
\left( \Ex[(\bg')^{\alpha z}] \right)^{1/\alpha}
\leq \left( ( \alpha z )! \Ex[ e^{\bg'}]
\right)^{1/\alpha}
\leq \left( ( \alpha z )! e^{1/2} \right)^{1/\alpha} \leq 2^{O(z)}.\]

Applying Markov's inequality now completes the proof.
\end{proof}

\subsection{Proof of Lemma~\ref{lem:approx-bound}}\label{sec:apr-bound-proof}

Let $\{ \hat{c}_i \}_{i \in [k]}$ be an optimal set of centers for the partition $(Y_1, \dots, Y_k)$ for the $(k,z)$-clustering problem on $\Pi(X)$, where $\hat{c}_i \in Y_i$.  Specifically, the points $\hat{c}_1,\dots, \hat{c}_k \in \R^t$ are those which minimize 
\[ \cost(\Pi(Y_1),\dots, \Pi(Y_k)) \eqdef \sum_{i=1}^k \sum_{x \in Y_i} \| \Pi(x) - \hat{c}_i \|_2^z. \]
To quantize the cost difference between the centers $c^*$ and the centers $
\hat{c}$
we analyze the following value. We assume we mapped every point of $X$ to $\Pi{c^*(X)}$, and we then compute the cost of the partition $Y_1,\dots, Y_k$ on this set. Formally, we let
\[ \val(\Pi, c^*, Y_1,\dots, Y_k) = \sum_{i=1}^k \sum_{j=1}^k |Y_i \cap X_j^*| \cdot \| \Pi(c_j^*) - \hat{c}_i \|_2^z. \]

First, we prove the following simple claim.
\begin{claim}\label{cl:center-set}
There exists a set of centers $c_1', \dots, c_k'$ (with possible repetitions) which are chosen among the points $\{ c_1^*,\dots, c_k^*\}$ such that
\begin{align*}
\sum_{i=1}^k \sum_{j=1}^k |Y_i \cap X_j^*| \cdot \| \Pi(c_j^*) - \Pi(c_i') \|_2^z \leq 2^{z} \cdot \val(\Pi, c^*, Y_1,\dots, Y_k).
\end{align*}
\end{claim}

\begin{proof}
We will prove the claim using the probabilistic method. For every $i \in [k]$, consider the distribution over center $\{ c_1^*,\dots ,c_k^*\}$ which samples a center $\bc_i'$ as
\begin{align*}
\Prx_{\bc_i'}\left[ \bc_i' = c_j^*\right] = \dfrac{|Y_i \cap X_j^*|}{\sum_{\ell=1}^d |Y_i \cap X_j^*|}.
\end{align*}
Then, we upper bound the expected cost of using the centers $\bc_i'$. Using Lemma~\ref{lem:jensen}, 
\begin{align*}
&\Ex\left[ \sum_{i=1}^k \sum_{j=1}^k |Y_i \cap X_j^*| \cdot \| \Pi(c_j^*) - \Pi(\bc_i') \|_2^z\right] \\
\leq {} & \sum_{i=1}^k \sum_{j=1}^k |Y_i \cap X_j^*| \cdot \Ex\left[\left( \| \Pi(c_j^*) - \hat{c}_i \|_2 + \| \Pi(\bc_i') - \hat{c}_i\|_2 \right)^z \right] \\
\leq {} & 2^{z-1} \sum_{i=1}^k \sum_{j=1}^k |Y_i \cap X_j^*| \cdot \| \Pi(c_j^*) - \hat{c}_i\|_2^z + 2^{z-1} \sum_{i=1}^k \left(\sum_{j=1}^k |Y_i \cap X_j^*| \right) \Ex\left[ \| \Pi(\bc_i') - \hat{c}_i\|_2^z\right] \\
= {} & 2^{z-1} \cdot \val(\Pi, c^*, Y_1,\dots, Y_k) + 2^{z-1} \sum_{i=1}^k \left( \sum_{j=1}^k |Y_i \cap X_j^*| \right) \sum_{\ell=1}^k \dfrac{|Y_i \cap X_{\ell}^*|}{\sum_{j=1}^k |Y_i \cap X_j^*|} \cdot \| \Pi(c_{\ell}^*) - \hat{c}_i\|_2^z \\
= {} & 2^{z} \cdot \val(\Pi, c^*, Y_1,\dots, Y_k).\qedhere
\end{align*}
\end{proof}

We now upper bound $\cost_z(Y_1,\dots, Y_k)$ in terms of $\cost_z(X_1^*,\dots, X_k^*)$. We do this by going through the centers chosen according to Claim~\ref{cl:center-set}. This will allow us to upper bound the cost of clustering with $(Y_1,\dots, Y_k)$ in terms of the $\cost_z(X_1^*,\dots, X_k^*)$ as well as clustering cost involving only pairwise distances from $\{ c_1^*,\dots, c_k^*\}$. Then, we relate to distances after applying the map $\Pi$. Specifically, first notice that if we consider the set of centers $c_1',\dots, c_k'$ chosen from Claim~\ref{cl:center-set}
\begin{align}
\cost_z(Y_1,\dots, Y_n) &\leq \sum_{i=1}^k \sum_{x \in Y_i} \| x - c_i' \|_2^z \nonumber \\
			&\leq \sum_{i=1}^k \sum_{j=1}^k \sum_{x \in Y_i \cap X_j^*} \left( \| x - c_j^*\|_2 + \| c_j^* - c_i'\|_2 \right)^z \nonumber \\
			&\leq 2^{z-1} \cdot \cost_z(X_1^*,\dots, X_k^*) + 2^{z-1} \sum_{i=1}^k \sum_{j=1}^k |Y_i \cap X_j^*| \cdot \| c_j^* - c_i' \|_2^z, \label{eq:haha1}
\end{align}
where the third inequality uses Lemma~\ref{lem:jensen} once more.
Note that the right-most summation of (\ref{eq:haha1}) involves distances which are only among $c_1^*,\dots, c_k^*$, so by the first assumption of the map $\Pi$ and Claim~\ref{cl:center-set}, we may upper bound
\begin{align}
\sum_{i=1}^k \sum_{j=1}^k |Y_i \cap X_j^*| \cdot \| c_j^* - c_i'\|_2^z &\leq \sfD_1^z \sum_{i=1}^k \sum_{j=1}^k |Y_i \cap X_j^*| \cdot \| \Pi(c_j^*) - \Pi(c_i') \|_2^z \nonumber \\
	&\leq \sfD_1^z \cdot 2^{z} \cdot \val(\Pi, c^*, Y_1,\dots, Y_k). \label{eq:haha2}
\end{align}
Combining (\ref{eq:haha1}) and (\ref{eq:haha2}), we may upper bound
\begin{align}
\cost_z(Y_1,\dots, Y_n) &\leq  2^{z-1} \cdot \cost_z(X_1^*,\dots, X_k^*) + \sfD_1^z \cdot 2^{2z-1} \cdot \val(\Pi, c^*, Y_1,\dots, Y_k) \nonumber \\
					&= 2^{z-1} \cdot \cost_z(X_1^*,\dots, X_k^*)  + \sfD_1^z \cdot 2^{2z-1} \sum_{i=1}^k \sum_{j=1}^k |Y_i \cap X_j^*| \cdot \| \Pi(c_j^*) - \hat{c}_i\|_2^z. \label{eq:haha3}
\end{align}
We continue upper bounding the right-most expression in (\ref{eq:haha3}) by applying the triangle inequality:
\begin{align}
&\sum_{i=1}^k \sum_{j=1}^k |Y_i \cap X_j^*| \cdot \| \Pi(c_j^*) - \hat{c}_i\|_2^z \nonumber \\
&\qquad \leq  2^{z-1} \sum_{i=1}^k \sum_{j=1}^k \sum_{x \in Y_i \cap X_j^*} \| \Pi(x) - \Pi(c_j^*)\|_2^z + 2^{z-1} \sum_{i=1}^k \sum_{j=1}^k \sum_{x \in Y_i \cap X_j^*} \| \Pi(x) - \hat{c}_i\|_2^z\nonumber \\
&\qquad  \leq 2^{z-1} \sfD_2 \cdot \cost_z(X_1^*,\dots, X_k^*) + 2^{z-1} \cdot \cost_z(\Pi(Y_1),\dots, \Pi(Y_k)). \label{eq:haha4}
\end{align}
By the third assumption of the lemma, we note that 
\begin{align}
\cost_z(\Pi(Y_1),\dots, \Pi(Y_k)) &\leq \sfD_3 \cdot \min_{c_1,\dots, c_k \in \R^t} \sum_{x \in X} \min_{j \in[k]} \| \Pi(x) - c_j\|_2^z \nonumber \\
&\leq \sfD_3 \cdot \sum_{j=1}^k \sum_{x \in X_j} \| \Pi(x) - \Pi(c_j^*)\|_2^z \leq \sfD_3 \sfD_2 \cdot \cost_z(X_1^*, \dots, X_k^*). \label{eq:haha5}
\end{align}
Summarizing by plugging (\ref{eq:haha4}) and (\ref{eq:haha5}) into (\ref{eq:haha3}), we can upper bound
\begin{align*}
\cost_z(Y_1,\dots, Y_n) \leq \left( 2^{z-1} + 2^{3z-2} \cdot \sfD_1^z \sfD_2 (1 + \sfD_3) \right) \cdot \cost_z(X_1^*,\dots, X_k^*).
\end{align*}

\section{Efficient Seeding in One Dimension}
\label{sec:runtime}
In this section, we prove \Cref{thm:runtime}, which shows an upper bound for the running time of our algorithm \name. As in \Cref{thm:apx-gen-z}, we consider a generalized version of \name where we run $k$-means++ seeding for general $z\ge 1$ (\Cref{def:k-means++}) in Step 2. We prove the following generalized version of \Cref{thm:runtime}:
\begin{theorem}[\Cref{thm:runtime} for general $z \ge 1$]
\label{thm:runtime-general}
Let $X = \{ x_1,\ldots,x_n\} \subset \R^d$ be a dataset consisting of $n$ points in $d$ dimensions. Assume that $d \le \nnz(X)$, which can be ensured after removing redundant dimensions $j\in [d]$ where the $j$-th coordinate of every $x_i$ is zero.
For any $z \ge 1$, the algorithm \name (for general $z$ as in \Cref{thm:apx-gen-z}) has expected running time $O(\nnz(X) + 2^{z/2}n\log n)$ on $X$. For any $\delta\in (0,1/2)$, with probability at least $1-\delta$, the algorithm runs in time $O(\nnz(X) + 2^{z/2}n\log (n/\delta))$. Moreover, the algorithm always runs in time $O(\nnz(X) + n\log n + nk)$.
\end{theorem}
To prove \Cref{thm:runtime}, we show an efficient implementation (\Cref{alg:1d-kmeans++}) of the $k$-means++ seeding procedure that runs in expected time $O(2^{z/2}n\log n)$ for one-dimensional points (\Cref{thm:seeding-runtime}). A naive implementation of the seeding procedure would take $\Theta(nk)$ time in one dimension because we need $\Theta(n)$ time to update $p_i$ and sample from the $D^2$ distribution to add each of the $k$ centers. To obtain an improved and provable running time, we use a basic binary tree data structure to sample from the $D^2$ distribution more efficiently, and we use a potential argument to bound the number of updates to $p_i$.

The data structure $S$ we use in
\Cref{alg:1d-kmeans++} can be implemented as a basic binary tree, as described in more detail in \Cref{sec:tree}. The data structure $S$ keeps track of $n$ nonnegative numbers $s_1,\ldots,s_n$ corresponding to $p_1^z,\ldots,p_n^z$ and it supports the following operations:
\begin{enumerate}
\item $\initialize(a)$. Given an array $a = (a_1,\ldots,a_n)\in \R_{\ge 0}^n$, the operation $\initialize(a)$ creates a data structure $S$ that keeps track of the numbers $s_1,\ldots,s_n$ initialized so that $(s_1,\ldots,s_n) = (a_1,\ldots,a_n)$. This operation runs in $O(n)$ time.
\item $\fsum(S)$. The operation $\fsum(S)$ returns the sum $s_1 + \cdots + s_n$. This operation runs in $O(1)$ time, as the value will be maintained as the data structure is updated.
\item $\find(S,r)$. Given a number $r\in [0,\sum_{i=1}^n s_i)$, the operation $\find(S,r)$ returns the unique index $\ell \in \{1,\ldots,n\}$ such that
\[
\sum_{i = 1}^{\ell -1}s_{i} \le r < \sum_{i=1}^\ell s_{i}.
\]
This operation runs in $O(\log n)$ time.
\item $\update(S,a,i_1,i_2)$. Given an array $a = (a_1,\ldots,a_n)\in \R_{\ge 0}^n$ and indices $i_1,i_2$ satisfying $1\le i_1 \le i_2 \le n$, the operation $\update(S,a,i_1,i_2)$ performs the updates 
$s_i\gets a_i$ for every $i = i_1,i_1+1,\ldots,i_2$. This operation runs in $O((i_2 - i_1 + 1) + \log n)$ time.
\end{enumerate}

\begin{algorithm}
\SetKwInput{Input}{Input}
\SetKwInput{Output}{Output}
\Input{Points $x_1,\ldots,x_n\in \R$; $k\in \Z$ satisfying $1\le k \le n$; real number $z\ge 1$.}
\Output{Centers $x_{\ell_1},\ldots,x_{\ell_k}\in \R$; assignment $\sigma:[n]\to [k]$.}
Sort and re-order the points so that $x_1 \le \cdots \le x_n$\;\label{line:1}
Choose $\ell_1$ uniformly at random from $\{1,\ldots,n\}$\;
Initialize $a = (a_1,\ldots,a_n)$ by setting $a_i \gets |x_i - x_{\ell_1}|^z$ for every $i = 1,\ldots,n$\;
$S\gets\initialize(a)$\;\label{line:before-for}
\For{$t = 2,\ldots,k$}{\label{line:for}
Choose $r$ uniformly at random from $[0,\fsum(S))$\label{line:for-1}\;
$\ell_t \gets \find(S,r)$;\quad $a_{\ell_t} \gets 0$;\quad $i\gets \ell_t - 1$; \quad $j\gets \ell_t + 1$\label{line:ell}\label{line:for-2}\;

\While{$i\ge 0$ \And $|x_i - x_{\ell_t}|^z < a_i$}{
	$a_i\gets |x_i - x_{\ell_t}|^z$\label{line:ai}\;
	$i\gets i - 1$\;
}
\While{$j\le n$ \And $|x_j - x_{\ell_t}|^z < a_j$}{
	$a_j\gets |x_j - x_{\ell_t}|^z$\;\label{line:aj}
	$j\gets j + 1$\;
}
\update($S,a,i + 1,j - 1$)\;
}
Sort and re-order $\ell_1,\ldots,\ell_k$ so that $\ell_1 \le \cdots \le \ell_k$ \label{line:assignment-1}\;
$i\gets 1$;\quad $j\gets 1$\;
\While (\tcc*[f]{Assign $x_i$ to the closest center $x_{\ell_{\sigma(i)}}$ among $x_{\ell_1},\ldots,x_{\ell_k}.$}){$i\le n$}
{
\eIf{$j < k$ \And $|x_i - x_{\ell_j}| \ge |x_i - x_{\ell_{j+1}}|$}
{$j\gets j + 1$\;}
{$\sigma(i)\gets j$\; $i\gets i+ 1$\;}
}
\Return $x_{\ell_1},\ldots,x_{\ell_k}$ and $\sigma$ (converted to the old ordering of $x_1,\ldots,x_n$ before Line~\ref{line:1})\label{line:return}\;
\caption{Efficient $k$-means++ seeding in one dimension}
\label{alg:1d-kmeans++}
\end{algorithm}
The following claim shows that \Cref{alg:1d-kmeans++} correctly implements the $k$-means++ seeding procedure in one dimension. 
\begin{claim}
\label{claim:correct}
Consider the values of $t, a_1,\ldots,a_n$ and the data structure $S$ at the beginning of each iteration of the for-loop (i.e., right before Line~\ref{line:for-1}). Let $s_1,\ldots,s_n$ be the numbers the data structure $S$ keeps track of. For every $i = 1,\ldots,n$, define $p_i:= \min_{t' = 1,\ldots,t - 1}|x_i - x_{\ell_{t'}}|$. 
Then $s_i = a_i = p_i^z$ for every $i = 1,\ldots,n$.
Consequently, the distribution of $\ell_t$ at \Cref{line:for-2} conditioned on the execution history so far satisfies $\Pr[\ell_t = i] = p_i^z/\sum_{i'=1}^np_{i'}^z$ for every $i = 1,\ldots,n$.
\end{claim}
The claim follows immediately by induction over the iterations of the for-loop based on the description of the data structure $S$ and its operations above.
The following lemma bounds the running time of \Cref{alg:1d-kmeans++}:
\begin{lemma}[\Cref{lm:runtime} for general $z \ge 1$]
\label{thm:seeding-runtime}
The expected running time of \Cref{alg:1d-kmeans++} is $O(2^{z/2}n \log n)$. For any $\delta\in (0,1/2)$, with probability at least $1-\delta$, \Cref{alg:1d-kmeans++} runs in time $O(2^{z/2}n\log (n/\delta))$. Moreover, \Cref{alg:1d-kmeans++} always runs in time $O(n \log n + nk)$.
\end{lemma}
Before proving \Cref{thm:seeding-runtime}, we first use it to
prove \Cref{thm:runtime-general}.
\begin{proof}[Proof of \Cref{thm:runtime-general}]
\Cref{thm:seeding-runtime} bounds the running time of Step 2 of our algorithm \name defined in \Cref{sec:overview}. Now we show that Step 1 (random one-dimensional projection) can be performed in time $O(\nnz(X) + n)$. Indeed, $x_i'$ can be computed as $x_i' = \sum_jx_{ij}v_j$ where the sum is over all the non-zero coordinates $x_{ij}$ of $x_i$, and each $v_j$ is drawn independently from the one-dimensional standard Gaussian (the value of $v_j$ should be shared for all $i$). The time needed to compute $x_1',\ldots,x_n'$ in this way is $O(\nnz(X) + n)$. In Step 3 of \name, we compute the center of mass for every cluster. This can be done in time $O(\nnz(X) + n)$ by summing up the points in each cluster and dividing each sum by the number of points in that cluster.
\end{proof}
The key step towards proving \Cref{thm:seeding-runtime} is to bound the number of updates to $a$ at \Cref{line:ai,line:aj}. As the algorithm starts by sorting the $n$ input points, which can be done in time $O(n \log n)$, we can assume that the points are sorted such that $x_1\le \cdots \le x_n$. For $i= 1,\ldots,n$ and $t = 2,\ldots,k$, we define $\xi(i,t) = 1$ if $a_i$ is updated at \Cref{line:ai} in iteration $t$ and define $\xi(i,t) = 0$ otherwise. Here, we denote each iteration of the for-loop beginning at \Cref{line:for} by the value of the iterate $t$. We define $u_i:=\sum_{t = 2}^k \xi(i,t)$ to be the number of times $a_i$ gets updated at \Cref{line:ai}. The following lemma gives upper bounds on $u_i$ both in expectation and with high probability:

\begin{lemma}
\label{lm:pi}
For every $i = 1,\ldots,n$, it holds that
\[
\E[u_i] \le O(2^{z/2}\log n).
\]
Moreover, for some absolute constant $B > 0$ and for every $\delta\in (0,1/2)$, it holds that
\[
\Pr[u_i \le B2^{z/2}\log(n/\delta)] \ge 1 - \delta.
\]
\end{lemma}
Before proving \Cref{lm:pi}, we first use it to prove \Cref{thm:seeding-runtime}.
\begin{proof}[Proof of \Cref{thm:seeding-runtime}]
Recall that for $i = 1,\ldots, n$ and $t = 2,\ldots, k$, we define $\xi(i,t) = 1$ if $a_i$ is updated at \Cref{line:ai} in iteration $t$ and define $\xi(i,t) = 0$ otherwise. Similarly, for $j = 1,\ldots,n$ and $t = 2,\ldots,k$, we define $\xi'(j,t)= 1$ if $a_j$ is updated at \Cref{line:aj} in iteration $t$ and define $\xi'(j,t) = 0$ otherwise.

The computation at Lines~\ref{line:1}-\ref{line:before-for} takes $O(n\log n)$ time. The computation at Lines~\ref{line:assignment-1}-\ref{line:return} takes $O(k\log k + n) = O(n\log n)$ time. For $t = 2,\ldots,k$, iteration $t$ of the for-loop takes time $O(\log n + \sum_{i=1}^n\xi(i,t) + \sum_{i=1}^n\xi'(i,t))$. Summing them up, the total running time of \Cref{alg:1d-kmeans++} is
\begin{equation}
\label{eq:runtime}
O\left(n\log n +  \sum_{i=1}^n\sum_{t = 2}^k\xi(i,t) + \sum_{i=1}^n\sum_{t = 2}^k\xi'(i,t)\right).
\end{equation}
By \Cref{lm:pi},
\begin{equation}
\label{eq:runtime-exp-1}
\E\left[\sum_{i=1}^n\sum_{t = 2}^k\xi(i,t)\right] = \E\left[\sum_{i=1}^n u_i\right] = O(2^{z/2}n \log n).
\end{equation}
Also, for any $\delta'\in (0,1/2)$, setting $\delta = \delta'/n$ in \Cref{lm:pi}, by the union bound we have
\begin{align}
\Pr\left[\sum_{i=1}^n\sum_{t = 2}^k\xi(i,t) \le 2B2^{z/2}n\log (n/\delta')\right] & \ge \Pr\left[\sum_{i=1}^n\sum_{t = 2}^k\xi(i,t) \le B2^{z/2}n\log (n/\delta)\right]\notag \\
& \ge 1 - n\delta\notag \\
& = 1 - \delta'.
\label{eq:runtime-prob-1}
\end{align}
Similarly to \eqref{eq:runtime-exp-1} and \eqref{eq:runtime-prob-1} we have
\begin{align}
\E\left[\sum_{i=1}^n\sum_{t = 2}^k\xi'(i,t)\right] & = O(2^{z/2}n \log n), \quad \text{and}\label{eq:runtime-exp-2}\\
\Pr\left[\sum_{i=1}^n\sum_{t = 2}^k\xi'(i,t) \le 2B2^{z/2}n\log (n/\delta')\right] & \ge 1 -\delta'.\label{eq:runtime-prob-2}
\end{align}
Plugging \eqref{eq:runtime-exp-1} and \eqref{eq:runtime-prob-1} into \eqref{eq:runtime} proves that the expected running time of \Cref{alg:1d-kmeans++} is $O(2^{z/2}n\log n)$. 
Choosing $\delta' = \delta/2$ for the $\delta$ in \Cref{thm:seeding-runtime} and plugging \eqref{eq:runtime-prob-1} and \eqref{eq:runtime-prob-2} into \eqref{eq:runtime}, we can use the union bound to conclude that with probability at least $1-\delta$ \Cref{alg:1d-kmeans++} runs in time $O(2^{z/2}n\log(n/\delta))$. Finally, plugging $\xi(i,t) \le 1$ and $\xi'(i,t) \le 1$ into \eqref{eq:runtime}, we get that \Cref{alg:1d-kmeans++} always runs in time $O(n\log n + nk)$.
\end{proof}

To prove \Cref{lm:pi}, for $i = 0,\ldots,n$ and $u = 0,\ldots,u_i$, we define $t(i,u)$ to be the smallest $t\in \{1,\ldots,k\}$ such that $\sum_{t' = 2}^{t}\xi(i,t') = u$. That is, $a_i$ gets updated at \Cref{line:ai} for the $u$-th time in iteration $t(i,u)$. Our definition implies that $t(i,0) = 1$ and $t(i,u)\in \{2,\ldots,k\}$ for $u = 1,\ldots,u_i$. We define a nonnegative potential function $\eta(i,u)$ as follows and show that it decreases exponentially in expectation as $u$ increases (\Cref{lm:eta}).

\begin{figure}[h]
\centering
\includegraphics[width=9cm]{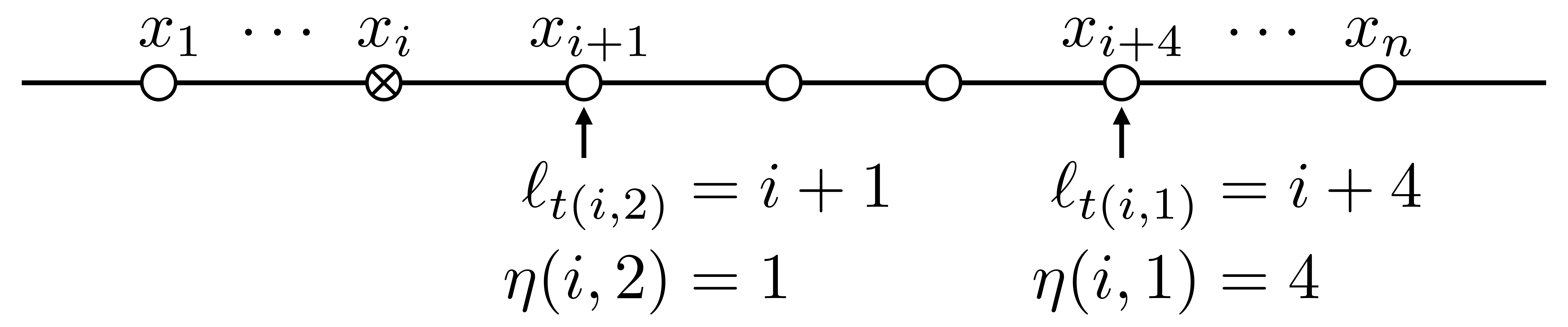}
\caption{An example illustrating the definition of the potential function $\eta$. Here, $a_i$ is updated at \Cref{line:ai} for the first time in iteration $t(i,1)$ when $\ell_{t(i,1)} = i+4$. Then $a_i$ gets updated at \Cref{line:ai} for the second time in iteration $t(i,2)$ when $\ell_{t(i,2)} = i + 1$. We always have $\ell_t > i$ whenever $a_i$ is updated at \Cref{line:ai}, and $\eta$ is the difference between $\ell_t$ and $i$. Thus in this example $\eta(i,1) = 4$ and $\eta(i,2) = 1$. If $a_i$ is never updated at \Cref{line:ai} after iteration $t(i,2)$, we define $\eta(i,u) = 0$ for $u = 3,4,\ldots$.}
\label{fig:eta}
\end{figure}

\paragraph{Potential Function $\eta(i,u)$.} For $t = 2,\ldots, k$, we consider the value of $\ell_t$ after \Cref{line:ell} in iteration $t$.
For $u = 1,\ldots, u_i$, we define $\eta(i,u)$ to be $\ell_{t(i,u)} - i$, which is guaranteed to be a positive integer by the definition of $t(i,u)$. 
Indeed, in the while-loop containing \Cref{line:ai}, $i$ starts from $\ell_t - 1$ and keeps decreasing, so whenever \Cref{line:ai} is executed, $i$ is smaller than $\ell_t$. In particular, $a_i$ is updated at \Cref{line:ai} in iteration $t(i,u)$ of the for-loop, so we have $i < \ell_{t(i,u)}$. We define $\eta(i,0) = n$, and for $u = u_i + 1, u_i + 2, \ldots,$ we define $\eta(i,u) = 0$. See \Cref{fig:eta} for an example illustrating the definition of $\eta(i,u)$.

\begin{lemma}[Potential function decrease]
\label{lm:eta}
For any $i\in \{1,\ldots,n\}$ and $u\in \Z_{\ge 0}$,
\[\E[\eta(i,u + 1)|\eta(i,0),\ldots,\eta(i,u)] \le \max\left\{0,\frac{2^{z/2}}{2^{z/2} + 1} \eta(i,u) - \frac 12\right\}.
\]
\end{lemma}
Before proving \Cref{lm:eta}, we first use it to prove \Cref{lm:pi}. Intuitively, \Cref{lm:eta} says that $\eta(i,u)$ decreases exponentially (in expectation) as a function of $u$. Since $\eta(i,u)$ is always a nonnegative integer, we should expect $\eta(i,u)$ to become zero as soon as $u$ exceeds a small threshold. Moreover, our definition ensures $\eta(i,u_i) > 0$, so $u_i$ must be smaller than the threshold. This allows us to show upper bounds for $u_i$ and prove \Cref{lm:pi}.
\begin{proof}[Proof of \Cref{lm:pi}]
Our definition of $\eta$ ensures $\eta(i,u_i) \ge 1$. By \Cref{lm:eta} and \Cref{lm:expected-stopping},
\[
\E[u_i + 1] \le \frac{\ln n}{\ln \frac{2^{z/2} + 1}{2^{z/2}}} + \frac{1}{1 - \frac{2^{z/2}}{2^{z/2} + 1}} = O(2^{z/2}\ln n) + 2^{z/2} + 1.
\]
This implies $\E[u_i] = O(2^{z/2}\log n)$. Moreover, by \Cref{lm:eta}, 
\[
\E[\eta(i,u)] \le \eta(i,0)\left(\frac{2^{z/2}}{2^{z/2} + 1}\right)^u = n\left(\frac{2^{z/2}}{2^{z/2} + 1}\right)^{u}, 
\]
and thus, by Markov's inequality,
\[
\Pr[u_i \ge u] = \Pr[\eta(i,u) \ge 1] \le n\left(\frac{2^{z/2}}{2^{z/2} + 1}\right)^u.
\]
For any $\delta\in (0,1/2)$, choosing $u = \ln(n/\delta)/\ln(\frac{2^{z/2} + 1}{2^{z/2}}) = O(2^{z/2}\log (n/\delta))$ in the inequality above gives $\Pr[u_i\ge u] \le \delta$.
\end{proof}

We need the following helper lemma to prove \Cref{lm:eta}.
\begin{lemma}
\label{lm:d-ell}
The following holds at the beginning of each iteration of the for-loop in \Cref{alg:1d-kmeans++}, i.e., right before \Cref{line:for-1} is executed. Choose an arbitrary $i = 1,\ldots,n$ and define 
\begin{equation}
\label{eq:L}
L:=\{i\}\cup\{\ell\in \Z:i < \ell \le n, |x_i - x_\ell|^z < a_i\}.
\end{equation} 
Then for $\ell,\ell'\in L$ satisfying $\ell < \ell'$, it holds that $a_{\ell'} \le 2^za_\ell$.
\end{lemma}
\begin{proof}
For $t = 2,\ldots,k$, at the beginning of iteration $t$, the values $\ell_1,\ldots,\ell_{t-1}$ have been determined.
For every $x\in \R$, let $\rho(x)$ denote the value among $x_{\ell_1},\ldots,x_{\ell_{t-1}}$ closest to $x$.
By \Cref{claim:correct}, $a_i = |x_i - \rho(x_i)|^z$ for every $i\in [n]$. Now for a fixed $i\in [n]$, define $L$ as in \eqref{eq:L} and consider $\ell,\ell'\in L$ satisfying $\ell < \ell'$. It is easy to see that $x_i \le \rho(x_\ell) \le x_{\ell'}$ cannot hold because otherwise $a_i \le |x_i - \rho(x_\ell)|^z \le |x_i - x_{\ell'}|^z < a_i$, a contradiction. For the same reason, the inequality $x_i \le \rho(x_{\ell'}) \le x_{\ell'}$ cannot hold. Thus there are only three possible orderings of $x_i,x_\ell,x_{\ell'},\rho(x_\ell),\rho(x_{\ell'})$:
\begin{enumerate}
\item $\rho(x_\ell) = \rho(x_{\ell'}) < x_i \le x_{\ell} \le x_{\ell'}$;
\item $\rho(x_\ell) < x_i \le x_\ell \le x_{\ell'} < \rho(x_{\ell'})$;
\item $x_i \le x_{\ell} \le x_{\ell'} < \rho(x_{\ell}) = \rho(x_{\ell'})$.
\end{enumerate}
In scenario 3, it is clear that $a_{\ell'} = |x_{\ell'} - \rho(x_{\ell'})|^z \le |x_\ell - \rho(x_\ell)|^z = a_\ell$. In the first two scenarios, for any $t' = 0,\ldots,t - 1$,
\[
|x_i - x_{\ell_{t'}}| \ge |x_\ell - x_{\ell_{t'}}| - |x_\ell - x_i| \ge |x_\ell - \rho(x_\ell)| - |x_\ell - x_i| = |x_i - \rho(x_\ell)|.
\]
This implies that 
$\rho(x_\ell)$ is the closest point to $x_i$ among $x_{\ell_1},\ldots,x_{\ell_{t-1}}$. Therefore, $a_i = |x_i - \rho(x_\ell)|^z$. Jensen's inequality ensures $((g + h)/2)^z \le (g^z + h^z)/2$ for any $g,h\ge 0$, which implies $(g + h)^z \le 2^{z-1}g^z + 2^{z-1}h^z$. Therefore,
\[
a_{\ell'} \le |x_{\ell'} - \rho(x_\ell)|^z \le 2^{z-1} |x_{\ell'} - x_i|^z + 2^{z-1}|x_i - \rho(x_\ell)|^z < 2^za_i,
\]
whereas
\[
a_\ell = |x_\ell - \rho(x_\ell)|^z \ge |x_i - \rho(x_\ell)|^z = a_i.
\]
Thus, we have $a_{\ell'} \le 2^za_\ell$ in all three scenarios.
\end{proof}

\begin{proof}[Proof of \Cref{lm:eta}]
Throughout the proof,
we fix $i\in \{1,\ldots,n\}$ and $u\in \Z_{\ge 0}$ so that they are deterministic numbers. \Cref{alg:1d-kmeans++} is a randomized algorithm, and when we run it, exactly one of the following four events happens, and we define a random variable $t^*$ accordingly:
\begin{enumerate}
\item Event $E_1$: $u_i < u$. That is, $a_i$ gets updated at \Cref{line:ai} for less than $u$ times. In this case we have $\eta(i,u+1) = 0$ by our definition of $\eta$, and we define $t^* = +\infty$.
\item Event $E_2$: $u_i = u$ and $i$ is never chosen as $\ell_t$ at \Cref{line:ell}. In this case we also have $\eta(i,u+1) = 0$, and we also define $t^* = +\infty$.
\item Event $E_3$: $u_i = u$ and there exists $t \in \{2,3,\ldots,k\}$ such that $i$ is chosen as $\ell_t$ at \Cref{line:ell} in iteration $t$. This $t$ must satisfy $t > t(i, u)$, as all updates to $a_i$ in Line~\ref{line:ai} must happen before $x_i$ is chosen as a center. We define $t^* = t$ in this case. Again, we have $\eta(i, u+1) = 0$ in this case.
\item Event $E_4$: $u_i > u$. We define $t^*:= t(i,u+1) > t(i,u)$ in this case.
\end{enumerate}
Define $E^*:= E_3\cup E_4$. Since $\eta(i,u + 1) = 0$ under $E_1$ and $E_2$, it suffices to prove that\footnote{We have $\eta(i,u+1) = 0$ also for $E_3$, so one can also simply choose $E^* = E_4$. Choosing $E^* = E_3 \cup E_4$ helps us get improved constants in our bound.}
\begin{equation}
\label{eq:eta-1}
\E[\eta(i,u+1)|\eta(i,0),\ldots,\eta(i,u),E^*] \le \frac{2^{z/2}}{2^{z/2} + 1} \eta(i,u) - \frac 12.
\end{equation}
By our definition, the random variable $t^*$ takes its value in $\{2,3,\ldots,k\}\cup \{+\infty\}$. Moreover, $t^* = +\infty$ if and only if $E^*$ does not happen. Therefore, to prove \eqref{eq:eta-1}, it suffices to prove the following for every $t_0 = 2,3,\ldots,k$:
\begin{equation}
\label{eq:eta-2}
\E[\eta(i,u+1)|\eta(i,0),\ldots,\eta(i,u),t^* = t_0] \le \frac{2^{z/2}}{2^{z/2} + 1} \eta(i,u) - \frac 12.
\end{equation}
Consider a fixed $t_0\in \{2,3,\ldots,k\}$.
For $t^* = t_0$ to happen, the following must hold during the execution of \Cref{alg:1d-kmeans++} before iteration $t_0$: $a_i$ has been updated at \Cref{line:ai} for exactly $u$ times, and $i$ has not been chosen as $\ell_t$ at \Cref{line:ell}. Thus, the rest of the proof assumes that the execution history $H$ of \Cref{alg:1d-kmeans++} before iteration $t_0$ satisfies this property. Now we know that the values $\eta(i,0),\ldots,\eta(i,u)$ are determined by the execution history $H$. Lines~\ref{line:for-1}-\ref{line:for-2} guarantee that the distribution of $\ell_{t_0}$ satisfies
\[
\Pr[\ell_{t_0} = \ell|H] = \frac{a_\ell}{\sum_{j = 1}^n a_j} \quad \text{for every }\ell = 1,\ldots,n,
\]
where we use the values $a_1,\ldots,a_n$ right before iteration $t_0$ is executed. Moreover, conditioned on $H$, we have $t^* = t_0$ if and only if $\ell_{t_0}\in L$, where
\[
L = \{i\}\cup \{\ell\in \Z: i < \ell \le n, |x_\ell - x_i|^z < a_i\}.
\]
Therefore, if we further condition on $t^* = t_0$, we have $\ell_{t_0}\in L$ and 
\begin{equation}
\label{eq:ell-dist}
\Pr[\ell_{t_0} = \ell|H, t^* = t_0] = \frac{a_\ell}{\sum_{j\in L}a_j} \quad \text{for every }\ell \in L.
\end{equation}
When $t^* = t_0$, we have $\eta(i,u+1) = \ell_{t_0} - i$. Therefore, to prove \eqref{eq:eta-2}, it suffices to show that
\begin{equation}
\label{eq:eta-3}
\E[\ell_{t_0} - i|H,t^* = t_0] \le \frac{2^{z/2}}{2^{z/2} + 1} \eta(i,u) - \frac 12. 
\end{equation}

It is clear that we can write $L$ as $L = \{i,i+1,\ldots,\ell^*\}$ for some integer $\ell^* \ge i$. If $u > 0$, \Cref{claim:correct} implies $a_i \le |x_i - x_{\ell_{t(i,u)}}|^z$, and thus $\ell^* < \ell_{t(i,u)}$ and $\ell^* - i \le \ell_{t(i,u)} - i = \eta(i,u)$. If $u = 0$, we have $\eta(i,u) = n$, so it also holds that $\ell^* - i \le \eta(i,u)$.
By \eqref{eq:ell-dist} and \Cref{lm:d-ell}, we can set $\gamma = 2^z$ in \Cref{lm:helper-rank-decrease} to get
\[
\E[\ell_{t_0} - i|H,t^* = t_0] \le \frac{2^{z/2}}{2^{z/2} + 1}(\ell^* - i) - \frac 12 \le \frac{2^{z/2}}{2^{z/2} + 1} \eta(i,u) - \frac 12. 
\]
This proves \eqref{eq:eta-3} and thus proves the lemma.
\end{proof}

\subsection{Helper Lemmas}
\begin{lemma}
\label{lm:expected-stopping}
Let $M \ge 1$ and $\lambda\in (0,1)$ be parameters.
Let $\alpha_0,\alpha_1,\ldots\in [0,+\infty)$ be random variables satisfying $\alpha_0 = M$ and $\E[\alpha_{i+1}|\alpha_1,\ldots,\alpha_i] \le \lambda \alpha_i$ for every $i = 0,1,\ldots$. Let $t \ge 0$ be the smallest integer satisfying $\alpha_t < 1$. Then
\[
\E[t] \le \frac{\ln M}{\ln(1/\lambda)} + \frac{1}{1-\lambda}.
\]
\end{lemma}
\begin{proof}
For every $j = 0,1,\ldots$, we define a random variable $t_j:= \min\{t,j\}$. By the monotone convergence theorem, it suffices to show that
\begin{equation}
\label{eq:induction-stopping}
\E[t_j] \le \frac{\ln M}{\ln(1/\lambda)} + \frac{1}{1-\lambda} \quad \text{for every }j = 0,1,\ldots.
\end{equation}
We prove \eqref{eq:induction-stopping} by induction on $j$. When $j = 0$, we have $t_j = 0$, and the inequality above holds trivially. We assume that \eqref{eq:induction-stopping} holds for an arbitrary $j \in\Z_{\ge 0}$ and show that it also holds with $j$ replaced by $j + 1$. We have 
\begin{equation}
\label{eq:stopping-total}
\E[t_{j+1}] = 1 + \E[t_{j+1} -1] = 1 + \E[\E[(t_{j+1} - 1)|\alpha_1]]. 
\end{equation}
By our definition of $t_{j+1}$, we have $t_{j+1} - 1 = \min\{t - 1, j\}$. Applying our induction hypothesis on the sequence $\alpha_1,\alpha_2,\ldots$, we have
\begin{equation}
\label{eq:stopping-1}
\E[(t_{j+1} - 1)|\alpha_1] = \E[\min\{t - 1, j\}|\alpha_1] \le f(\alpha_1),
\end{equation}
where
\[
f(\alpha) = \begin{cases}
\frac{\alpha}{1-\lambda},& \text{if }\alpha\in [0,1);\\
\frac{\ln \alpha}{\ln(1/\lambda)} + \frac{1}{1-\lambda},& \text{if }\alpha \ge 1.
\end{cases}
\]
It is easy to check that $f$ is an increasing concave function of $\alpha\in [0,+\infty)$ and $1 + f(\lambda \alpha)\le f(\alpha)$ holds for every $\alpha \ge 1$.
Plugging \eqref{eq:stopping-1} into \eqref{eq:stopping-total}, we have
\[
\E[t_{j+1}] \le 1 + \E[f(\alpha_1)] \le 1 + f(\E[\alpha_1]) \le 1 + f(\lambda M) \le f(M) = \frac{\ln M}{\ln (1/\lambda)} + \frac 1{1-\lambda}.\qedhere
\]
\end{proof}
\begin{lemma}
\label{lm:helper-rank-decrease}
Let $\gamma\ge 1$ be a real number.
Let $\beta_0,\ldots,\beta_{m - 1}$ be non-negative real numbers such that for every $i,j\in \{0,\ldots,m - 1\}$ satisfying $i\le j$, it holds that $\beta_j \le \gamma \beta_i$. Then,
\[
\sum_{i=0}^{m - 1} i\beta_i \le \left(\frac{\sqrt \gamma \cdot m}{\sqrt \gamma + 1} -\frac 12\right)\sum_{i=0}^{m - 1} \beta_i.
\]
\end{lemma}
\begin{proof}
The lemma holds trivially if $\beta_0 = 0$ because  in this case $\beta_j \le \gamma \beta_0= 0$ for every $j = 0,\ldots,m - 1$. We thus assume w.l.o.g.\ that $\beta_0 > 0$. Define $\tau$ to be the unique real number satisfying
\[
\tau \sum_{i=0}^{m - 1} \beta_i - \sum_{i=0}^{m - 1} i\beta_i = 0.
\]
It is clear that $\tau \in [0,m - 1]$. Our goal is to prove that
\begin{equation}
\label{eq:goal-rank-decrease}
\tau \le \frac{\sqrt \gamma \cdot  m}{\sqrt \gamma  + 1} -\frac 12.
\end{equation}
Define $\beta_*:= \min_{0\le i \le \tau} \beta_i$. For every $i = 0,\ldots,m - 1$, we have $\beta_i \ge \beta_*$ if $i \le \tau$, and $\beta_i \le \gamma \beta_*$ if $i > \tau$. Therefore, defining $s:= \lfloor \tau \rfloor$, we have
\begin{align}
0 & = \tau \sum_{i=0}^{m - 1} \beta_i - \sum_{i=0}^{m - 1} i\beta_i\notag \\
& = \sum_{i=0}^{m - 1}(\tau - i)\beta_i\notag \\
& \ge \sum_{i \le \tau}(\tau - i)\beta_* + \sum_{i > \tau}(\tau - i)\gamma \beta_*\label{eq:helper-rank-decrease-1}\\
& = \frac{(s + 1)(2\tau - s)}2\cdot \beta_* + \frac{(m - s - 1)(2\tau - m - s)}2\cdot \gamma \beta_*.\label{eq:helper-rank-decrease-2}
\end{align}
Now, we show that $\beta_* > 0$. For the sake of contradiction, assume $\beta_* = 0$. We already assumed that $\beta_0 > 0$, so $\beta_* \ne \beta_0$. By the definition of $\beta_*$, this means that $\tau > 0$ and inequality \eqref{eq:helper-rank-decrease-1} is strict, leading to the false claim of 
\[
0 > \sum_{i \le \tau}(\tau - i)\beta_* + \sum_{i > \tau}(\tau - i)\gamma \beta_* = 0.
\]
Therefore, $\beta_* > 0$ must hold. Now we know that \eqref{eq:helper-rank-decrease-2} implies
\[
(s + 1)(2\tau - s) + (m - s - 1)(2\tau - m - s)\gamma  \le 0.
\]
Treating $s$ as a real-valued variable, the left-hand side is minimized when $s = \tau - 1/2$, giving us
\[
(\tau + 1/2)^2 - (m - \tau - 1/2)^2\gamma  \le 0.
\]
The inequality above implies
\[
(\tau + 1/2)^2 \le  (m - \tau - 1/2)^2\gamma.
\]
Taking square root for both sides and solving for $\tau$ gives \eqref{eq:goal-rank-decrease}.
\end{proof}

\section{Data Structure for Fast Sampling in Seeding}
\label{sec:tree}
In \Cref{sec:runtime}, our \Cref{alg:1d-kmeans++} uses a binary tree data structure $S$ that keeps track of $n$ nonnegative numbers $s_1,\ldots,s_n$ and supports several operations. Here, we describe the implementation of this data structure.
We assume that $n = 2^q$ for some nonnegative integer $q$. This is without loss of generality because we can choose $n'$ to be the number that satisfy $n\le n' < 2n$ and $n' = 2^q$ for some $q\in \Z_{\ge 0}$ and consider $s_1,\ldots,s_{n},s_{n+1},\ldots,s_{n'}$ with $s_{n+1} = \cdots = s_{n'} = 0$.
Under this assumption, the data structure $S$ is a complete binary tree with $q+1$ layers indexed by $0,\ldots,q$. In each layer $\zeta = 0,\ldots,q$ there are $2^\zeta$ nodes each corresponding to a set of indices from $\{1,\ldots,n\}$. The root, denoted by $v\sps 0_1$, is the unique node in layer $0$ and it corresponds to the entire set $V\sps 0_1:=\{1,\ldots,n\}$. For $\zeta = 0,\ldots, q - 1$, each node $v\sps \zeta_j$ in the $\zeta$-th layer has two children $v\sps {\zeta+1}_{2j-1},v\sps {\zeta+1}_{2j}$ in the $(\zeta+1)$-th layer corresponding to the sets $V\sps {\zeta+1}_{2j-1},V\sps {\zeta+1}_{2j}$, respectively, where $V\sps {\zeta+1}_{2j-1}$ is the smaller half of $V\sps \zeta_j$ and $V\sps {\zeta+1}_{2j}$ is the larger half. Thus,
\[
V\sps \zeta_j = \{i\in \Z:(j - 1)2^{q-\zeta} < i \le j2^{q - \zeta}\}.
\]
Each node $v\sps \zeta_j$ in the tree stores a sum $s\sps \zeta_j:=\sum_{i\in V\sps \zeta _j}s_i$.

\begin{figure}[h]
\centering
\includegraphics[width=9cm]{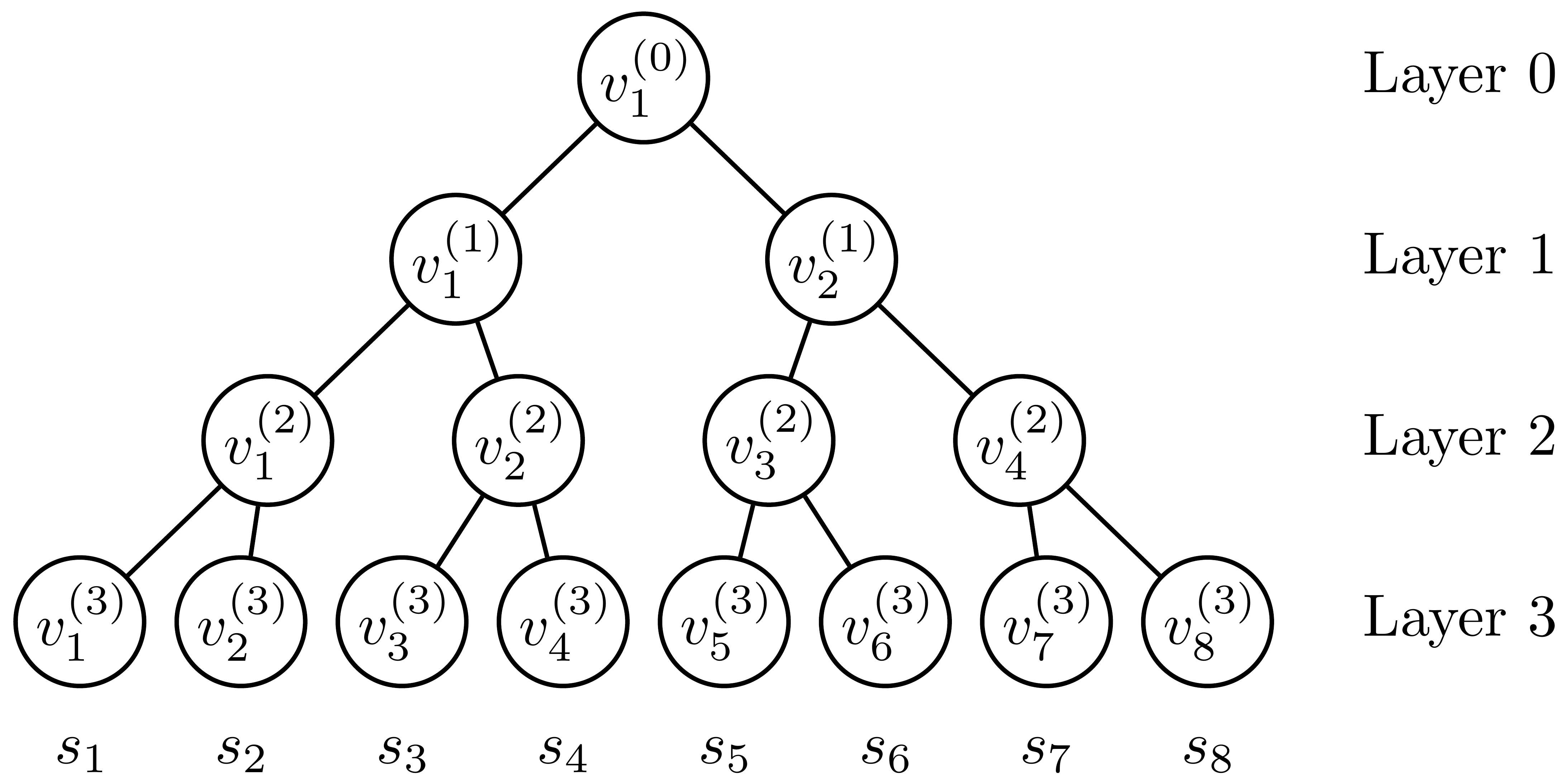}
\caption{An example of a basic binary tree data structure with $q = 3$.}
\label{fig:tree}
\end{figure}

The data structure supports the four types of operations needed in \Cref{sec:runtime} as follows:
\begin{enumerate}
\item $\initialize(a)$. Recursively run $\initialize$ on the first half $(a_1,\ldots,a_{n/2})$ and the second half $(a_{n/2+1},\ldots,a_n)$ to obtain the two subtrees rooted at $v\sps 1_1$ and $v\sps 1_2$. Then add a root $v\sps 0_1$ that stores $s\sps 0_1 \gets s\sps 1_1 + s\sps 1_2$.
\item $\fsum(S)$. Simply output $s\sps 0_1$.
\item $\find(S,r)$. If $r < s\sps 1_1$, recursively call $\find$ on the left subtree rooted at $v\sps 1_1$. Otherwise, recursively call $\find$ on the right subtree rooted at $v\sps 1_2$ with $r$ replaced by $r - s\sps 1_1$. Once we reach a leaf $v\sps q_\ell$, return $\ell$.
\item $\update(S,a,i_1,i_2)$. If $i_2 \le n/2$, recursively call $\update$ on the left subtree rooted at $v\sps 1_1$. If $i_1 > n/2$, recursively call $\update$ on the right subtree rooted at $v\sps 1_2$. Otherwise, we have $i_1 \le n/2 < i_2$ and we call $\update$ on the left subtree with indices $i_1, n/2$ and call $\update$ on the right subtree with indices $n/2 + 1, i_2$. In all cases, we update $s\sps 0_1\gets s\sps 1_1 + s\sps 1_2$ as the final step. The running time is proportional to the number of nodes we update. We need to update $s\sps \zeta_j$ stored at $v\sps \zeta_j$ only if $V\sps\zeta _j\cap\{i_1,\ldots,i_2\}\ne \emptyset$. For each $\zeta$, the number of such $j$ is at most $(i_2 - i_1 + 1)/2^{q - \zeta} + 2$. Summing up over $\zeta = 0,\ldots,q$,  the total number of nodes we need to update is $O((i_2 - i_1 + 1) + q) = O((i_2 - i_1 + 1) + \log n)$.
\end{enumerate}

\section{Experimental Results}\label{sec:experiments}

In this section, we outline the experimental evaluation of our algorithm. The experiments evaluate the algorithms in two different ways. For each, we measure the running time and the $k$-means cost of the resulting solution (the sum of squares of point-to-center-assigned distances). 
(1) First, we evaluate our algorithm as part of a pipeline incorporating a coreset construction 
-- the expected use case for our algorithm. 
(2) Second, we evaluate our algorithm by itself for approximate k-means clustering and compare it to k-means++~\cite{MR2485254}. As per Theorems~\ref{thm:runtime} and~\ref{thm:apx-ratio}, we expect our algorithm to be much faster but output an assignment of higher cost. Our goal is to quantify these differences empirically.

All experiments were run on Linux using a notebook with a 3.9 GHz 12th generation Intel Core i7 six-core processor and 32 GiB of RAM. All algorithms were implemented in C++, using the \texttt{blaze} library for matrix and vector operations performed on the dataset unless specified differently below. The code is publicly available on GitHub\footnote{\label{fn:github}\href{https://github.com/boredoms/prone}{\name GitHub repository: \texttt{https://github.com/boredoms/prone}}}.

For our experiments, we use the following four datasets:
\begin{enumerate}
\item\textbf{KDD}~\cite{kdd-cup}: Training data for the 2004 KDD challenge on protein homology. The dataset consists of $145751$ observations with $77$ real-valued features.
    
\item\textbf{Song}~\cite{bertin2011million}: Timbre information for $515345$ songs with $90$ features each, used for year prediction.
    
\item\textbf{Census}~\cite{Dua:2019}: 1990 US census data with $2458285$ observations, each with $68$ categorical features.

\item\textbf{Gaussian}: A synthetic dataset consisting of $240005$ points of dimension $4$. The points are generated by placing a standard normal distribution at a large positive distance from the origin on each axis and sampling $30000$ points. The points are then mirrored so the center of mass remains at the origin. Finally, 5 points are placed on the origin. This 
is an adversarial example for lightweight coresets~\cite{10.1145/3219819.3219973}, which are unlikely to sample points close to the mean of the dataset.
\end{enumerate}

\subsection{Coreset Construction Comparison} 

\paragraph{Experimental Setup.} Coreset constructions (with multiplicative approximation guarantees) always proceed by first finding an approximate clustering, which constitutes the bulk of the work. The approximate clustering defines a ``sensitivity sampling distribution'' (we expand on this in \cref{sec:prelims}, see also~\cite{bachem2017practical}), and a coreset is constructed by repeatedly sampling from the sensitivity sampling distribution. In our first experiment, we evaluate the choice of initial approximation algorithm used to define the sensitivity sampling distribution. We compare the use of $k$-means++ and \name. In addition, we also compare the lightweight coresets of \cite{10.1145/3219819.3219973}, which uses the distance to the center of mass as an approximation of the sensitivity sampling distribution. 

\begin{figure}
    \centering
    \includegraphics[width=0.85\textwidth]{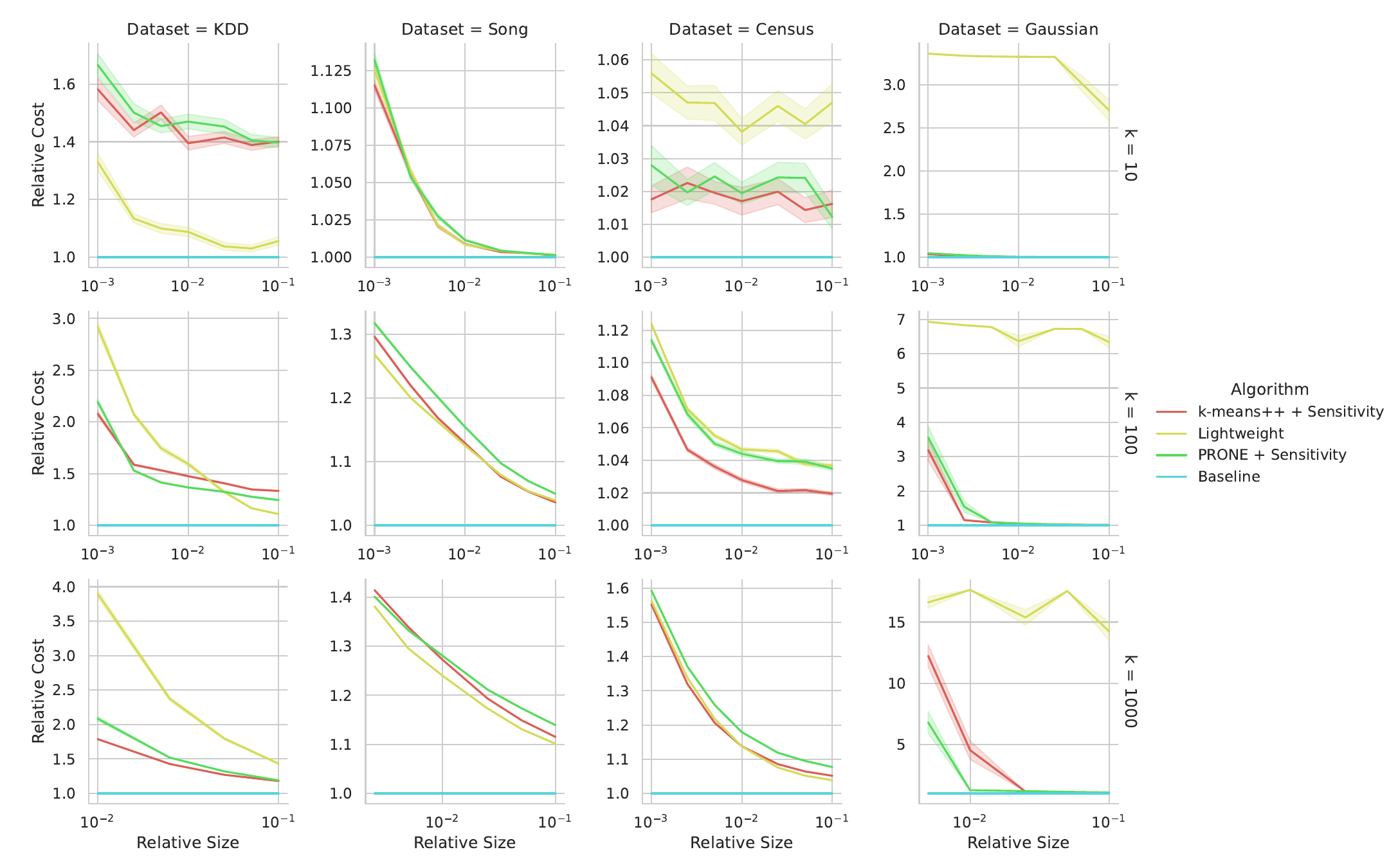}
    \caption{Plot of relative cost versus coreset size on our four datasets. 
    The shaded region indicates standard error. There is no data for relative size values where the coreset size is less than $k$.}
    \label{fig:coreset_qualy}
\end{figure}

For the remainder of this section, we refer to sensitivity sampling using k-means++ as \emph{Sensitivity} and lightweight coresets as \emph{Lightweight}. All three algorithms produce a coreset, and the experiment will measure the running time of the three algorithms (Table~\ref{tab:runtime}) and the quality of the resulting coresets (Figure~\ref{fig:coreset_qualy}). 

Once a coreset is constructed for each of the algorithms, we evaluate the quality of the coreset by computing the cost of the centers found when clustering the coreset (see Definition~\ref{def:k-z-cluster}).
We run a state-of-the-art implementation of Lloyd's $k$-means algorithm from the \texttt{scikit-learn} library~\cite{scikit-learn} with the default configuration (repeating 15 times and reporting the mean cost to reduce the variance). 
The resulting quality of the coresets is compared to a (computationally expensive) \emph{baseline}, which runs \texttt{k-means++} from the \texttt{scikit-learn} library, 
followed by Lloyd's algorithm with the default configuration on the entire dataset (repeated 5 times to reduce variance). 

We evaluate various choices of $k$ ($\{ 10, 100, 1000\}$) as well as coresets at various relative sizes, $\{0.001, 0.0025, 0.005, 0.01, 0.025, 0.05, 0.1\}$ times the size of the dataset. 
We use as performance metrics (1) a relative cost, which measures the average cost of the k-means solutions returned by Lloyd's algorithm on each coreset divided by the baseline, 
and (2) the running time of the coreset construction algorithm.

\paragraph{Results on Coreset Constructions.}
\emph{Relative cost.} Figure~\ref{fig:coreset_qualy} shows the coreset size ($x$-axis) versus the relative cost ($y$-axis). Each ``row'' of Figure~\ref{fig:coreset_qualy} corresponds to a different value for $k \in  \{10, 100, 1000\}$, and each ``column'' corresponds to a different dataset. Recall that the first three datasets (i.e., the first three columns) are real-world datasets, and the fourth column is the synthetic Gaussian dataset. We note our observations below:
\begin{itemize}
\item As expected, on all real-world data sets and all settings of $k$, the relative cost decreases as the coreset size increases. 
\item In real-world datasets, the specific relative cost of each coreset construction (\emph{Senstivity}, \emph{Lightweight}, and ours) depends on the dataset\footnote{The spike in relative cost for algorithm \emph{Sensitivity} on the KDD data set for relative size $5 \cdot 10^{-3}$ is due to outliers.}, but roughly speaking, all three share a similar trend. Ours and \emph{Sensitivity} are very close and never more than twice the baseline (usually much better). 
\item The big difference, distinguishing ours and \emph{Sensitivity} from \emph{Lightweight}, is the fourth column, the synthetic Gaussian dataset. For all settings of $k$, as the coreset size increases, \emph{Lightweight} exhibits a minimal cost decrease and is a factor of 2.7-17x times worse than ours and \emph{Sensitivity} (as well as the baseline). This is expected, as we constructed the synthetic Gaussian dataset to have arbitrarily high cost with \emph{Lightweight}. Due to its multiplicative approximation guarantee, our algorithm does not suffer this degradation. In that sense, our algorithm is more ``robust,'' and achieves worst-case multiplicative approximation guarantees for all datasets.
\end{itemize}

\emph{Running time.} In (the first table in) Table~\ref{tab:runtime},
we show the running time of the coreset construction algorithms as $k$ increases. 
Notice that as $k$ increases, the relative speedup of our algorithm and \emph{Lightweight} increases in comparison to \emph{Sensitivity}. This is because our algorithm and \emph{Lightweight} have running time which \emph{does not grow with $k$}. In contrast, the running time of 
\emph{Sensitivity} grows linearly in $k$. In summary, our coreset construction is between \textbf{33-192x} faster than \emph{Sensitivity} for large $k$.  
In addition, our algorithm runs about 3-5x slower than \emph{Lightweight}, depending on the dataset. Our analysis also shows this; both algorithms make an initial pass over the dataset, using $O(nd)$ time,  but ours uses an additional $O(n \log n)$ time to process.

\subsection{Direct k-Means++ Comparison}\label{sec:direct-comparison}

\paragraph{Experimental Setup.} This experiment compares our algorithm and $k$-means++ as a stand-alone clustering algorithm, as opposed to as part of a coreset pipeline. We implemented three variants of our algorithm. Each differs in how we sample the random one-dimensional projection. The first is a one-dimensional projection onto a standard Gaussian vector (zero mean and identity covariance). 
This approach risks collapsing an ``important'' feature, i.e.~a feature with high variance. To mitigate this, we implemented two {\em data-dependent} variants that use the variance, resp.~covariance of the data. Specifically, 
in the ``variance'' variant, we use a diagonal covariance matrix, where each entry in the diagonal is set to the empirical variance of the dataset along the corresponding feature. In the ``covariance'' variant, we use the empirical covariance matrix of the dataset. These variants aim to project along vectors that capture more of the variance of the data than when sampling a vector uniformly at random. Intuitively, the vectors sampled by the biased variants are more correlated with the first principal component of the dataset. 
For each of our algorithms, we evaluate the $k$-means cost of the output set $C$ of centers when assigning points to the closest center ($\cost_2(X,C)$ in Definition~\ref{def:k-z-cluster}) and when using our algorithm's assignment ($\cost_2(X,C,\sigma)$ defined in \eqref{eq:assignment-cost}). 

We evaluated the algorithms for every $k$ in $\{10, 25, 50, 100, 250, 500, 1000, 2500, 5000\}$ and $z = 2$, for solving $k$-means with the $\ell_2$-metric.
When evaluating the assignment cost, we ran each of our algorithms 100 times for each $k$ and five times when computing the nearest neighbor assignment, and we report the average cost of the solutions and the average running time. Due to lower variance and much higher runtime, k-means++ was run five times.

\begin{figure}
    \centering
    \includegraphics[width=0.9\textwidth]{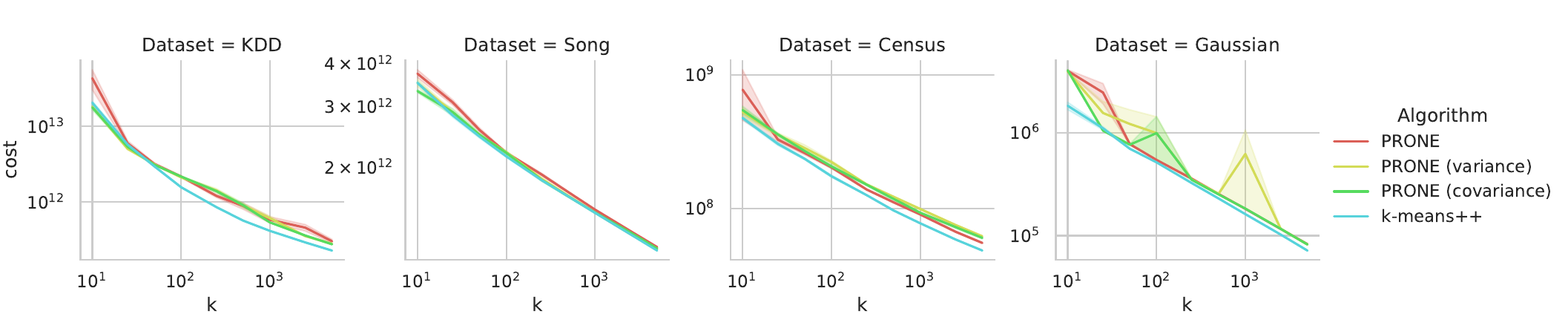}
    \includegraphics[width=0.9\textwidth]{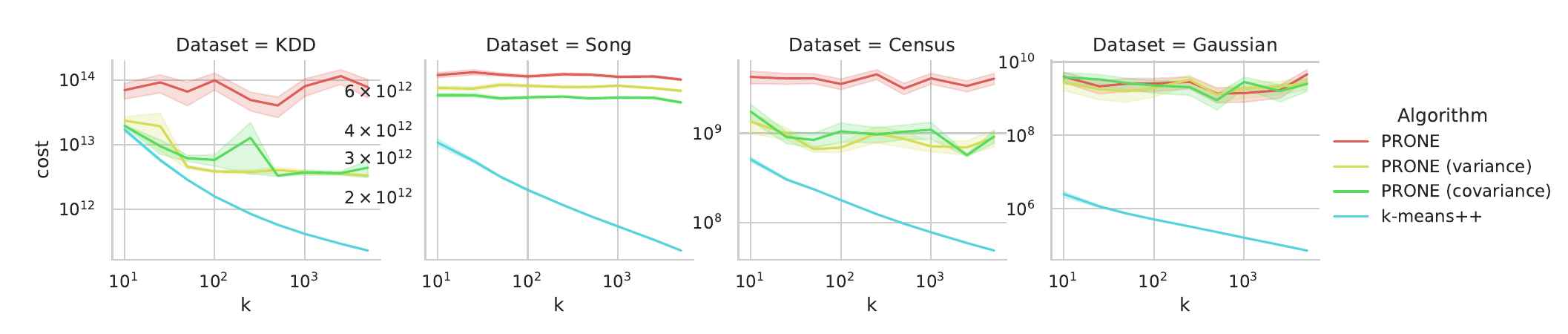}
    \caption{Clustering cost of all our variants compared to $k$-means++. The top row shows the $k$-means cost, and the bottom row shows the cost of the assignment produced by our algorithm.}
    \label{fig:cluster-cost}
\end{figure}

\begin{table}[ht]
    \centering
\scalebox{0.7}{
\begin{tabular}{llrrr}
\toprule
          & k & 10   &  100  &   1000 \\
Dataset & Algorithm &      &       &        \\
\midrule
Census & Lightweight &  7.3 &  69.4 &  670.2 \\
          & \name coreset &  1.5 &  14.1 &  136.3 \\
          & Sensitivity &  1.0 &   1.0 &    1.0 \\
Song & Lightweight &  8.9 &  87.2 &  875.3 \\
          & \name coreset &  2.1 &  19.9 &  187.9 \\
          & Sensitivity &  1.0 &   1.0 &    1.0 \\
KDD & Lightweight &  6.4 &  63.0 &  642.8 \\
          & \name coreset &  2.1 &  19.6 &  192.6 \\
          & Sensitivity &  1.0 &   1.0 &    1.0 \\
Gaussian & Lightweight &  2.4 &  17.6 &  174.8 \\
          & \name coreset &  0.5 &   3.8 &   33.7 \\
          & Sensitivity &  1.0 &   1.0 &    1.0 \\
\bottomrule
\end{tabular}
\quad
\begin{tabular}{llrrr}
\toprule
          & k   & 50   &   500  &    5000 \\
Dataset & Algorithm &      &      &     \\
\midrule
Census & \name &   7.5 &   73.2 &  662.5 \\
          & \name (variance) &    2.2 &     22.2 &   214.7 \\
          & \name (covariance) &  1.1 &     10.7 &    117.4 \\
          & k-means++  & 1.0 &    1.0 &    1.0 \\
Song & \name &   9.7 &    95.5 &   837.5 \\
          & \name (variance) &  2.3 &    23.1 &    217.2 \\
          & \name (covariance) &   0.8 &    8.2 &   82.4 \\
          & k-means++ &    1.0 &    1.0 &    1.0 \\
KDD & \name &   6.9 &    68.3 &   727.5 \\
          & \name (variance) &  3.1 &   32.0 &   312.4 \\
          & \name (covariance) &  1.3 &    12.9 &   128.4 \\
          & k-means++ &  1.0 &  1.0 &  1.0  \\
Gaussian & \name &   1.9 &  18.3 &  165.9 \\
          & \name (variance) &  2.0 &   17.7 &   162.9 \\
          & \name (covariance)  &  1.7 &    16.1 &  152.6 \\
          & k-means++ &  1.0 &  1.0 &  1.0 \\
\bottomrule
\end{tabular}
}
    \caption{Average speedup over sensitivity sampling across all relative sizes for constructing coresets (in the first table) and average speedup over $k$-means++  as a stand-alone clustering algorithm (in the second table). The tables with the full range of parameters can be found in the appendix.}
    \label{tab:runtime}
\end{table}

\paragraph{Results on Direct $k$-Means++ Comparison.}
\emph{Cost.} 
Figure~\ref{fig:cluster-cost} (on top)
shows the cost of the centers found by our algorithm compared to those found by the k-means++ algorithm after computing 
\emph{the optimal assignment of points to the centers} 
(computing this 
takes time $O(ndk)$). 
That is, we compare the values of $\cost_2(X,C)$ in Definition~\ref{def:k-z-cluster}.
In summary, the k-means cost of all three variants of our algorithm are roughly the same and closely match that of k-means++. On the Gaussian data set, one run of the biased algorithm failed to pick a center from the cluster at the origin, leading to a high ``outlier'' cost and a corresponding spike in the plot.

We also compared the k-means cost for the assignment \emph{computed by our algorithm} (so that our algorithm only takes time $O(nd + n \log n)$ and \emph{not} $O(ndk)$) with the cost of k-means++ (bottom row of Figure~\ref{fig:cluster-cost}).
That is, we compare the values of $\cost_2(X,C,\sigma)$ defined in \eqref{eq:assignment-cost}.
The clustering cost of our algorithms is higher than that of k-means++. This is the predicted outcome from our theoretical results; recall Theorem~\ref{thm:apx-ratio} gives a $\poly(k)$-approximation, as opposed to $O(\log k)$ from $k$-means++. 

On the real-world data sets, it is between one order of magnitude (for $k=10$) and two orders of magnitude (for $k=5000$) worse than k-means++ for our unbiased variant and between a factor 2 (for $k=10$) and one order of magnitude (for $k=5000$) worse than k-means++ for our biased and covariance variants.

\emph{Running time.} Table~\ref{tab:runtime} shows the relative running time of our algorithm compared to k-means++, assuming that no nearest-center assignment is computed. Our algorithms are 
designed to have a running time independent of $k$, so we can see, from the second table in Figure~\ref{tab:runtime}, all of our variants offer significant speedups.
\begin{itemize}
    \item The running time of our algorithm stays almost constant as $k$ increases while the running time of k-means++ scales linearly with $k$. Specifically for $k=25$, even our slowest variants have about the same running time as k-means++, while for $k=5000$, it is at least \textbf{82x} faster, and our fastest version is up to \textbf{837x} faster over k-means++.
    \item The two variants can affect the quality of the chosen centers by up to an order of magnitude, but they are also significantly slower. The ``variance'' and ``covariance'' variants are slower (between 2-4x slower and up to 10x slower, respectively) than the standard variant, and they also become slower as the dimensionality $d$ increases. We believe these methods could be further sped up, as the \texttt{blaze} library's variance computation routine appears inefficient for our use case.
\end{itemize}

\subsection{Improved Approximation Ratio}

\paragraph{Experimental Setup. } This experiment aims to compare the algorithmic approach outlined in Theorem~\ref{thm:improved-apx} to the direct use of \name as a clustering algorithm as was done in Section~\ref{sec:direct-comparison}. For this, we use \name as the approximation algorithm for sensitivity sampling and then cluster the coreset using a weighted variant of the $k$-means++ algorithm. This approach is termed \name (boosted) in the rest of this section. This pipeline requires as parameters the number of centers $k$ and a hyperparameter $\alpha$ indicating the size of the coreset produced by sensitivity sampling. We aim to compare the clustering cost (see Definition~\ref{def:k-z-cluster}) and running time of our approach to that of $k$-means++.

We run both algorithms on the datasets described in Section~\ref{sec:experiments} and choose $k \in \{ 10, 25, 50, 100, 250,\\ 500, 1000, 2500, 5000 \}$ and $\alpha \in \{0.001, 0.01, 0.1 \}$. Each algorithm is run $5$ times.

\begin{figure}
    \centering
    \includegraphics[width=0.9\textwidth]{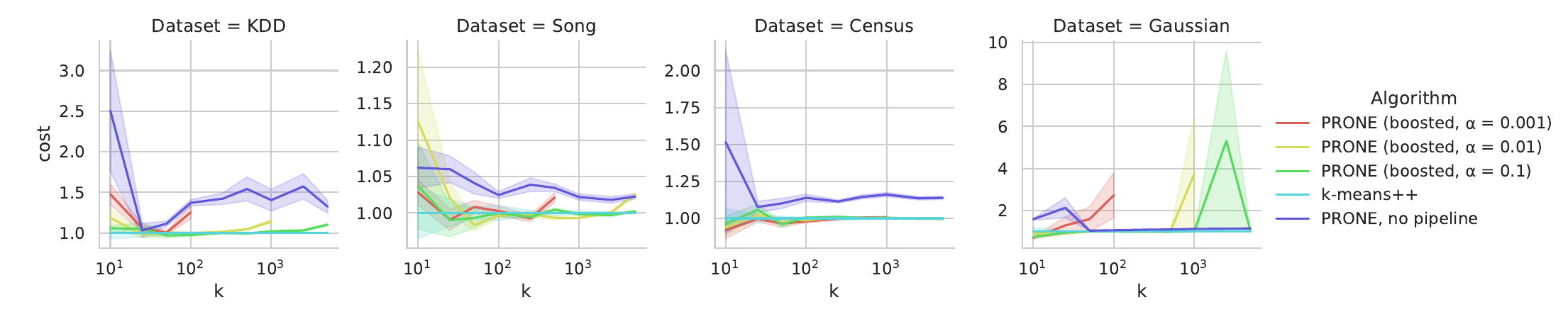}
    \caption{Clustering cost of the boosted variants compared to $k$-means++. Lines in the plot show the cost of centers produced by the boosted algorithm relative to $k$-means++ for centers ranging from 10 to 5000. Dark blue indicates the non-boosted version.}
    \label{fig:cluster-cost-boosted}
\end{figure}

\paragraph{Results on Improved Approximation Ratio}
\emph{Costs. } Figure~\ref{fig:cluster-cost-boosted} shows the costs of centers produced by this algorithm relative to the cost of centers produced by $k$-means++. It also contains the $(k, 2)$-clustering costs of \name relative to $k$-means++. We can see that on all real datasets, \name (boosted) produces solutions of the same or better quality than $k$-means++, as long as $\alpha n \ll k$. This shows that although \name by itself produces centers of worse quality, the \name (boosted) variant produces centers of the same quality as vanilla $k$-means++.
When $\alpha n \approx k$, we observe an uptick in cost before the end of the lines corresponding to $\alpha \in \{0.1, 0.01\}$ in the plots for KDD, Song, and Gaussian. The boosted approach outperforms \name, which is usually worse by a constant factor compared to the other algorithms, and it helps to reduce significantly the amount of variance in the quality of solutions. On the Gaussian dataset, we observed a failure to sample a point from the central cluster, which explains the spike at $k = 2500$ for the line corresponding to $\alpha = 0.1$.

\emph{Running time. } Table~\ref{tab:boosted-running-time} shows the speedup of the boosted approach versus using plain $k$-means++, for the time taken to compute the centers. The running time of our algorithms now scales with $k$, but at a slower rate compared to $k$-means++, as we have to run it on a much smaller dataset. Once again, we observe significant speedups, especially as $k$ grows.
\begin{itemize}
    \item As expected, the speedup depends on the choice of the hyperparameter $\alpha$. We observe diminishing returns for larger $\alpha$ as $k$ scales, with the speedup remaining mostly constant for $k \ge 100$ across datasets, except for Gaussian. This is because the algorithm's running time is dominated by the time it takes to execute $k$-means++ on the coreset, which has $O(ndk)$ asymptotic running time. The speedups we can achieve using this method are significant, up to 118x faster than $k$-means++. We expect that on massive datasets, even greater speedups can be achieved.

    \item Interestingly, the speedup can come very close to or even out scale $\alpha$, as observed on the KDD and Song datasets. The final stage of the boosted approach executes $k$-means++ on a coreset of size $\alpha n$, so the running time of this step should be $O(\alpha n dk)$. The observed additional speedup may be due to better cache and memory utilization in the $k$-means++ step of the algorithm. 
\end{itemize}

\begin{table}[ht]
    \centering
    \scalebox{0.85}{
\begin{tabular}{llrrrrrrrrr}
\toprule
          & Centers & 10   & 25   & 50   &  100  &  250  &  500  &   1000 &   2500 &   5000 \\
Dataset & Algorithm &      &      &      &       &       &       &        &        &        \\
\midrule
Census & \name (boosted, $\alpha = 0.001$) & 1.6 & 4.0 & 7.9 & 15.6 & 36.8 & 69.5 & 118.5 & - & - \\
          & \name (boosted, $\alpha = 0.01$) & 1.5 & 3.7 & 6.7 & 11.7 & 20.8 & 28.5 & 30.1 & 36.2 & 41.9 \\
          & \name (boosted, $\alpha = 0.1$) & 1.0 & 1.8 & 2.3 & 2.7 & 3.0 & 3.2 & 3.0 & 3.0 & 3.2 \\
          & $k$-means++ &  1.0 &  1.0 &  1.0 &   1.0 &   1.0 &   1.0 &    1.0 &    1.0 &    1.0 \\
Song & \name (boosted, $\alpha = 0.001$) & 2.1 & 5.3 & 10.7 & 21.3 & 51.7 & 97.5 & - & - & - \\
          & \name (boosted, $\alpha = 0.01$) & 2.1 & 5.0 & 9.8 & 18.5 & 38.0 & 59.1 & 81.5 & 108.3 & 117.0 \\
          & \name (boosted, $\alpha = 0.1$) & 1.3 & 2.2 & 2.8 & 3.4 & 3.8 & 4.0 & 4.0 & 4.0 & 3.9 \\
          & $k$-means++ &  1.0 &  1.0 &  1.0 &   1.0 &   1.0 &   1.0 &    1.0 &    1.0 &    1.0 \\
KDD & \name (boosted, $\alpha = 0.001$) & 1.8 & 5.0 & 9.6 & 20.6 & - & - & - & - & - \\
          & \name (boosted, $\alpha = 0.01$) & 1.9 & 4.8 & 9.0 & 17.5 & 37.1 & 59.2 & 92.5 & - & - \\
          & \name (boosted, $\alpha = 0.1$) & 1.4 & 2.7 & 3.6 & 4.9 & 5.7 & 6.2 & 7.1 & 6.6 & 6.4 \\
          & $k$-means++ &  1.0 &  1.0 &  1.0 &   1.0 &   1.0 &   1.0 &    1.0 &    1.0 &    1.0 \\
Gaussian & \name (boosted, $\alpha = 0.001$) & 0.5 & 0.9 & 1.9 & 3.6 & - & - & - & - & - \\
          & \name (boosted, $\alpha = 0.01$) & 0.5 & 1.0 & 2.0 & 3.4 & 8.2 & 14.8 & 25.6 & - & - \\
          & \name (boosted, $\alpha = 0.1$) & 0.4 & 0.8 & 1.4 & 2.2 & 4.0 & 5.0 & 6.1 & 6.8 & 7.2 \\
          & $k$-means++ &  1.0 &  1.0 &  1.0 &   1.0 &   1.0 &   1.0 &    1.0 &    1.0 &    1.0 \\
\bottomrule
\end{tabular}
}
    \caption{Average speedup when computing a clustering and assignment for different datasets relative to k-means++. In other words, each cell contains $T_{k\text{-means++}} / T_\text{PRONE}$. Missing entries denote the case of $\alpha n > k$.}
    \label{tab:boosted-running-time}
\end{table}

\section{Conclusion and Limitations} 

To summarize, we present a simple algorithm that provides a new tradeoff between running time and approximation ratio. Our algorithm runs in expected time $O(\nnz(X) + n\log n)$ to produce a $\poly(k)$-approximation; with additional $\poly(kd) \cdot \log n$ time, we improve the approximation to $O(\log k)$. This latter bound matches that of $k$-means++ but offers a significant speedup.

Within a pipeline for constructing coresets, our experiments show that the quality of the coreset produced (when using our algorithm as the initial approximation) outperforms the sensitivity sampling algorithm. It is slower than the lightweight coreset algorithm, but it is more ``robust'' as it is independent of the diameter of the data set.
It does not suffer from the drawback of having an additive error linear in the diameter of the dataset, which can arbitrarily increase the cost of the lightweight coreset algorithm. 
When computing an optimal assignment for the centers returned by our algorithm, its cost roughly matches the cost for k-means++. When directly using the assignment produced by one variant of our algorithm, its cost is between a factor 2 and 10 worse while being up to 300 times faster.

Our experiments and running time analysis show that our algorithm is very efficient. However, the clustering quality achieved by our algorithm is sometimes not as good as other, slower algorithms. We show that this limitation is insignificant when we use our algorithm to construct coresets. It remains an interesting open problem to understand the best clustering quality (e.g., in terms of approximation ratio) an algorithm can achieve while being as efficient as ours, i.e., running in time $O(nd + n\log n)$. Another interesting problem is whether other means of projecting the dataset into a $O(1)$ dimensional space exist, which lead to algorithms with improved approximation guarantees and running time faster than $O(ndk)$.

\section*{Acknowledgements}
Moses Charikar was supported by a Simons Investigator award.
Lunjia Hu was supported by Moses Charikar’s and Omer Reingold’s Simons Investigators awards, Omer Reingold’s NSF Award IIS-1908774, and the Simons Foundation Collaboration on the Theory of Algorithmic Fairness. Part of this work was done while Erik Waingarten was a postdoc at Stanford University, supported by an NSF postdoctoral fellowship and by Moses Charikar’s Simons Investigator Award.

\erclogowrapped{5\baselineskip}This project has received funding from the European Research Council (ERC) under the European Union's Horizon 2020 research and innovation programme (Grant agreement No. 101019564 ``The Design of Modern Fully Dynamic Data Structures (MoDynStruct)'' and the Austrian Science Fund (FWF) project 
Z 422-N, project
“Static and Dynamic Hierarchical Graph Decompositions”, I 5982-N, and project “Fast Algorithms for a Reactive Network Layer (ReactNet)”, P 33775-N, with additional funding from the netidee SCIENCE Stiftung, 2020–2024.

\bibliographystyle{alpha}
\bibliography{arxiv-version}

\newcommand{\etalchar}[1]{$^{#1}$}
\begin{thebibliography}{CALNF{\etalchar{+}}20}

\bibitem[ADHP09]{ADHP09}
Daniel Aloise, Amit Deshpande, Pierre Hansen, and Preyas Popat.
\newblock {NP}-hardness of {E}uclidean sum-of-squares clustering.
\newblock {\em Machine Learning}, 75(2):245--248, 2009.

\bibitem[AHPV{\etalchar{+}}05]{AHV05}
Pankaj~K Agarwal, Sariel Har-Peled, Kasturi~R Varadarajan, et~al.
\newblock Geometric approximation via coresets.
\newblock {\em Combinatorial and computational geometry}, 52(1):1--30, 2005.

\bibitem[AIR18]{AIR18}
Alexandr Andoni, Piotr Indyk, and Ilya Razenshteyn.
\newblock Approximate nearest neighbor search in high dimensions.
\newblock In {\em Proceedings of the International Congress of Mathematicians:
  Rio de Janeiro 2018}, pages 3287--3318. World Scientific, 2018.

\bibitem[ANFSW17]{ANSW17}
Sara Ahmadian, Ashkan Norouzi-Fard, Ola Svensson, and Justin Ward.
\newblock Better guarantees for k-means and {E}uclidean k-median by primal-dual
  algorithms.
\newblock In {\em 2017 IEEE 58th Annual Symposium on Foundations of Computer
  Science (FOCS)}, pages 61--72, 2017.

\bibitem[ARR98]{ARR98}
Sanjeev Arora, Prabhakar Raghavan, and Satish Rao.
\newblock Approximation schemes for {E}uclidean k-medians and related problems.
\newblock In {\em Proceedings of the Thirtieth Annual ACM Symposium on Theory
  of Computing}, STOC '98, page 106–113, New York, NY, USA, 1998. Association
  for Computing Machinery.

\bibitem[AV07]{MR2485254}
David Arthur and Sergei Vassilvitskii.
\newblock {\tt k-means++}: the advantages of careful seeding.
\newblock In {\em Proceedings of the {E}ighteenth {A}nnual {ACM}-{SIAM}
  {S}ymposium on {D}iscrete {A}lgorithms}, pages 1027--1035. ACM, New York,
  2007.

\bibitem[BBCA{\etalchar{+}}19]{BBCGS19}
Luca Becchetti, Marc Bury, Vincent Cohen-Addad, Fabrizio Grandoni, and Chris
  Schwiegelshohn.
\newblock Oblivious dimension reduction for k-means: Beyond subspaces and the
  johnson-lindenstrauss lemma.
\newblock In {\em Proceedings of the 51st Annual ACM SIGACT Symposium on Theory
  of Computing}, STOC 2019, page 1039–1050, New York, NY, USA, 2019.
  Association for Computing Machinery.

\bibitem[BBK16]{pmlr-v48-bottesch16}
Thomas Bottesch, Thomas Bühler, and Markus Kächele.
\newblock Speeding up k-means by approximating {E}uclidean distances via block
  vectors.
\newblock In Maria~Florina Balcan and Kilian~Q. Weinberger, editors, {\em
  Proceedings of The 33rd International Conference on Machine Learning},
  volume~48 of {\em Proceedings of Machine Learning Research}, pages
  2578--2586, New York, New York, USA, 20--22 Jun 2016. PMLR.

\bibitem[BFL16]{DBLP:journals/corr/BravermanFL16}
Vladimir Braverman, Dan Feldman, and Harry Lang.
\newblock New frameworks for offline and streaming coreset constructions.
\newblock {\em CoRR}, abs/1612.00889, 2016.

\bibitem[BLHK16a]{NIPS2016_d67d8ab4}
Olivier Bachem, Mario Lucic, Hamed Hassani, and Andreas Krause.
\newblock Fast and provably good seedings for k-means.
\newblock In D.~Lee, M.~Sugiyama, U.~Luxburg, I.~Guyon, and R.~Garnett,
  editors, {\em Advances in Neural Information Processing Systems}, volume~29.
  Curran Associates, Inc., 2016.

\bibitem[BLHK16b]{Bachem_Lucic_Hassani_Krause_2016}
Olivier Bachem, Mario Lucic, S.~Hamed Hassani, and Andreas Krause.
\newblock Approximate k-means++ in sublinear time.
\newblock {\em Proceedings of the AAAI Conference on Artificial Intelligence},
  30(1), Feb. 2016.

\bibitem[BLK17]{bachem2017practical}
Olivier Bachem, Mario Lucic, and Andreas Krause.
\newblock Practical coreset constructions for machine learning.
\newblock {\em arXiv preprint arXiv:1703.06476}, 2017.

\bibitem[BLK18]{10.1145/3219819.3219973}
Olivier Bachem, Mario Lucic, and Andreas Krause.
\newblock Scalable k -means clustering via lightweight coresets.
\newblock In {\em Proceedings of the 24th ACM SIGKDD International Conference
  on Knowledge Discovery \& Data Mining}, KDD '18, page 1119–1127, New York,
  NY, USA, 2018. Association for Computing Machinery.

\bibitem[BMEWL11]{bertin2011million}
Thierry Bertin-Mahieux, Daniel~P.W. Ellis, Brian Whitman, and Paul Lamere.
\newblock The million song dataset.
\newblock In {\em {Proceedings of the 12th International Conference on Music
  Information Retrieval ({ISMIR} 2011)}}, 2011.

\bibitem[BPR{\etalchar{+}}15]{BPRST15}
Jaros\l{}aw Byrka, Thomas Pensyl, Bartosz Rybicki, Aravind Srinivasan, and Khoa
  Trinh.
\newblock An improved approximation for k-median, and positive correlation in
  budgeted optimization.
\newblock In {\em Proceedings of the Twenty-Sixth Annual ACM-SIAM Symposium on
  Discrete Algorithms}, SODA '15, page 737–756, USA, 2015. Society for
  Industrial and Applied Mathematics.

\bibitem[Bri]{FAISSMANUAL}
James Briggs.
\newblock Faiss: The missing manual.
\newblock \url{https://www.pinecone.io/learn/faiss/}.

\bibitem[BZMD14]{boutsidis2014randomized}
Christos Boutsidis, Anastasios Zouzias, Michael~W Mahoney, and Petros Drineas.
\newblock Randomized dimensionality reduction for $ k $-means clustering.
\newblock {\em IEEE Transactions on Information Theory}, 61(2):1045--1062,
  2014.

\bibitem[CA18]{C18}
Vincent Cohen-Addad.
\newblock A fast approximation scheme for low-dimensional k-means.
\newblock In {\em Proceedings of the Twenty-Ninth Annual ACM-SIAM Symposium on
  Discrete Algorithms}, SODA '18, page 430–440, USA, 2018. Society for
  Industrial and Applied Mathematics.

\bibitem[CAFS21]{10.1145/3477541}
Vincent Cohen-Addad, Andreas~Emil Feldmann, and David Saulpic.
\newblock Near-linear time approximation schemes for clustering in doubling
  metrics.
\newblock {\em J. ACM}, 68(6), oct 2021.

\bibitem[CAKM16]{CKM16}
Vincent Cohen-Addad, Philip~N. Klein, and Claire Mathieu.
\newblock Local search yields approximation schemes for k-means and k-median in
  {E}uclidean and minor-free metrics.
\newblock In {\em 2016 IEEE 57th Annual Symposium on Foundations of Computer
  Science (FOCS)}, pages 353--364, 2016.

\bibitem[CALNF{\etalchar{+}}20]{NEURIPS2020_babcff88}
Vincent Cohen-Addad, Silvio Lattanzi, Ashkan Norouzi-Fard, Christian Sohler,
  and Ola Svensson.
\newblock Fast and accurate k-means++ via rejection sampling.
\newblock In H.~Larochelle, M.~Ranzato, R.~Hadsell, M.F. Balcan, and H.~Lin,
  editors, {\em Advances in Neural Information Processing Systems}, volume~33,
  pages 16235--16245. Curran Associates, Inc., 2020.

\bibitem[CALSS22]{CLSS22}
Vincent Cohen-Addad, Kasper~Green Larsen, David Saulpic, and Chris
  Schwiegelshohn.
\newblock Towards optimal lower bounds for k-median and k-means coresets.
\newblock In {\em Proceedings of the 54th Annual ACM SIGACT Symposium on Theory
  of Computing}, pages 1038--1051, 2022.

\bibitem[CASS21]{CSS21}
Vincent Cohen-Addad, David Saulpic, and Chris Schwiegelshohn.
\newblock A new coreset framework for clustering.
\newblock In {\em Proceedings of the 53rd Annual ACM SIGACT Symposium on Theory
  of Computing}, pages 169--182, 2021.

\bibitem[CEM{\etalchar{+}}15]{CEMMP15}
Michael~B. Cohen, Sam Elder, Cameron Musco, Christopher Musco, and Madalina
  Persu.
\newblock Dimensionality reduction for k-means clustering and low rank
  approximation.
\newblock In {\em Proceedings of the Forty-Seventh Annual ACM Symposium on
  Theory of Computing}, STOC '15, page 163–172, New York, NY, USA, 2015.
  Association for Computing Machinery.

\bibitem[CGPR20]{pmlr-v119-choo20a}
Davin Choo, Christoph Grunau, Julian Portmann, and Vaclav Rozhon.
\newblock k-means++: few more steps yield constant approximation.
\newblock In Hal~Daumé III and Aarti Singh, editors, {\em Proceedings of the
  37th International Conference on Machine Learning}, volume 119 of {\em
  Proceedings of Machine Learning Research}, pages 1909--1917. PMLR, 13--18 Jul
  2020.

\bibitem[Che09]{doi:10.1137/070699007}
Ke~Chen.
\newblock On coresets for k-median and k-means clustering in metric and
  {E}uclidean spaces and their applications.
\newblock {\em SIAM Journal on Computing}, 39(3):923--947, 2009.

\bibitem[CPL18]{capo2018efficient}
Marco Cap{\'o}, Aritz P{\'e}rez, and Jose~A Lozano.
\newblock An efficient k-means clustering algorithm for massive data.
\newblock {\em arXiv preprint arXiv:1801.02949}, 2018.

\bibitem[Cur17]{curtin2017dual}
Ryan~R Curtin.
\newblock A dual-tree algorithm for fast k-means clustering with large k.
\newblock In {\em Proceedings of the 2017 SIAM International Conference on Data
  Mining}, pages 300--308. SIAM, 2017.

\bibitem[DFK{\etalchar{+}}04]{DFKVV04}
Petros Drineas, Alan Frieze, Ravi Kannan, Santosh Vempala, and Vishwanathan
  Vinay.
\newblock Clustering large graphs via the singular value decomposition.
\newblock {\em Machine learning}, 56:9--33, 2004.

\bibitem[DG17]{Dua:2019}
Dheeru Dua and Casey Graff.
\newblock {UCI} machine learning repository, 2017.

\bibitem[Dra13]{drake2013faster}
Jonathan Drake.
\newblock {\em Faster k-means clustering.}
\newblock PhD thesis, Baylor University, 2013.

\bibitem[DZS{\etalchar{+}}15]{pmlr-v37-ding15}
Yufei Ding, Yue Zhao, Xipeng Shen, Madanlal Musuvathi, and Todd Mytkowicz.
\newblock Yinyang k-means: A drop-in replacement of the classic k-means with
  consistent speedup.
\newblock In Francis Bach and David Blei, editors, {\em Proceedings of the 32nd
  International Conference on Machine Learning}, volume~37 of {\em Proceedings
  of Machine Learning Research}, pages 579--587, Lille, France, 07--09 Jul
  2015. PMLR.

\bibitem[EKSX96]{DBSCAN}
Martin Ester, Hans-Peter Kriegel, J\"{o}rg Sander, and Xiaowei Xu.
\newblock A density-based algorithm for discovering clusters in large spatial
  databases with noise.
\newblock In {\em Proceedings of the Second International Conference on
  Knowledge Discovery and Data Mining}, KDD'96, page 226–231. AAAI Press,
  1996.

\bibitem[Elk03]{10.5555/3041838.3041857}
Charles Elkan.
\newblock Using the triangle inequality to accelerate k-means.
\newblock In {\em Proceedings of the Twentieth International Conference on
  International Conference on Machine Learning}, ICML'03, page 147–153. AAAI
  Press, 2003.

\bibitem[Fel20]{F20}
Dan Feldman.
\newblock Introduction to core-sets: an updated survey.
\newblock {\em arXiv preprint arXiv:2011.09384}, 2020.

\bibitem[FL11]{FL11}
Dan Feldman and Michael Langberg.
\newblock A unified framework for approximating and clustering data.
\newblock In {\em Proceedings of the forty-third annual ACM symposium on Theory
  of computing}, pages 569--578, 2011.

\bibitem[FRS16]{FRS16}
Zachary Friggstad, Mohsen Rezapour, and Mohammad~R. Salavatipour.
\newblock Local search yields a ptas for k-means in doubling metrics.
\newblock In {\em 2016 IEEE 57th Annual Symposium on Foundations of Computer
  Science (FOCS)}, pages 365--374, 2016.

\bibitem[FSS20]{FSS20}
Dan Feldman, Melanie Schmidt, and Christian Sohler.
\newblock Turning big data into tiny data: Constant-size coresets for k-means,
  pca, and projective clustering.
\newblock {\em SIAM Journal on Computing}, 49(3):601--657, 2020.

\bibitem[GOR{\etalchar{+}}22]{GORSV22}
Fabrizio Grandoni, Rafail Ostrovsky, Yuval Rabani, Leonard~J. Schulman, and
  Rakesh Venkat.
\newblock A refined approximation for {E}uclidean k-means.
\newblock {\em Inf. Process. Lett.}, 176(C), jun 2022.

\bibitem[Ham10]{hamerly2010making}
Greg Hamerly.
\newblock Making k-means even faster.
\newblock In {\em Proceedings of the 2010 SIAM international conference on data
  mining}, pages 130--140. SIAM, 2010.

\bibitem[HCLM09]{10.14778/1687627.1687771}
Oktie Hassanzadeh, Fei Chiang, Hyun~Chul Lee, and Ren\'{e}e~J. Miller.
\newblock Framework for evaluating clustering algorithms in duplicate
  detection.
\newblock {\em Proc. VLDB Endow.}, 2(1):1282–1293, aug 2009.

\bibitem[HPM04]{10.1145/1007352.1007400}
Sariel Har-Peled and Soham Mazumdar.
\newblock On coresets for k-means and k-median clustering.
\newblock In {\em Proceedings of the Thirty-Sixth Annual ACM Symposium on
  Theory of Computing}, STOC '04, page 291–300, New York, NY, USA, 2004.
  Association for Computing Machinery.

\bibitem[JDS10]{jegou2010product}
Herve Jegou, Matthijs Douze, and Cordelia Schmid.
\newblock Product quantization for nearest neighbor search.
\newblock {\em IEEE transactions on pattern analysis and machine intelligence},
  33(1):117--128, 2010.

\bibitem[JJ19]{DBSCAN++}
Jennifer Jang and Heinrich Jiang.
\newblock Dbscan++: Towards fast and scalable density clustering.
\newblock In {\em International conference on machine learning}, pages
  3019--3029. PMLR, 2019.

\bibitem[KC04]{kdd-cup}
2004 KDD~Cup.
\newblock Protein homology dataset.
\newblock {\em available at https://osmot.cs.cornell.edu/kddcup/datasets.html},
  2004.

\bibitem[KMN{\etalchar{+}}00]{10.1145/336154.336189}
Tapas Kanungo, David~M. Mount, Nathan~S. Netanyahu, Christine Piatko, Ruth
  Silverman, and Angela~Y. Wu.
\newblock The analysis of a simple k-means clustering algorithm.
\newblock In {\em Proceedings of the Sixteenth Annual Symposium on
  Computational Geometry}, SCG '00, page 100–109, New York, NY, USA, 2000.
  Association for Computing Machinery.

\bibitem[KMN{\etalchar{+}}02]{1017616}
T.~Kanungo, D.M. Mount, N.S. Netanyahu, C.D. Piatko, R.~Silverman, and A.Y. Wu.
\newblock An efficient k-means clustering algorithm: analysis and
  implementation.
\newblock {\em IEEE Transactions on Pattern Analysis and Machine Intelligence},
  24(7):881--892, 2002.

\bibitem[KR99]{KR99}
Stavros~G. Kolliopoulos and Satish Rao.
\newblock A nearly linear-time approximation scheme for the {E}uclidean
  k-median problem.
\newblock In Jaroslav Ne{\v{s}}et{\v{r}}il, editor, {\em Algorithms - ESA' 99},
  pages 378--389, Berlin, Heidelberg, 1999. Springer Berlin Heidelberg.

\bibitem[Llo82]{1056489}
Stuart~P. Lloyd.
\newblock Least squares quantization in {PCM}.
\newblock {\em IEEE Transactions on Information Theory}, 28(2):129--137, 1982.

\bibitem[LS13]{10.1145/2488608.2488723}
Shi Li and Ola Svensson.
\newblock Approximating k-median via pseudo-approximation.
\newblock In {\em Proceedings of the Forty-Fifth Annual ACM Symposium on Theory
  of Computing}, STOC '13, page 901–910, New York, NY, USA, 2013. Association
  for Computing Machinery.

\bibitem[LS19]{pmlr-v97-lattanzi19a}
Silvio Lattanzi and Christian Sohler.
\newblock A better k-means++ algorithm via local search.
\newblock In Kamalika Chaudhuri and Ruslan Salakhutdinov, editors, {\em
  Proceedings of the 36th International Conference on Machine Learning},
  volume~97 of {\em Proceedings of Machine Learning Research}, pages
  3662--3671. PMLR, 09--15 Jun 2019.

\bibitem[LST17]{LST17}
Weiwei Liu, Xiaobo Shen, and Ivor Tsang.
\newblock Sparse embedded $ k $-means clustering.
\newblock {\em Advances in neural information processing systems}, 30, 2017.

\bibitem[MMR19]{MMR19}
Konstantin Makarychev, Yury Makarychev, and Ilya Razenshteyn.
\newblock Performance of johnson-lindenstrauss transform for k-means and
  k-medians clustering.
\newblock In {\em Proceedings of the 51st Annual ACM SIGACT Symposium on Theory
  of Computing}, STOC 2019, page 1027–1038, New York, NY, USA, 2019.
  Association for Computing Machinery.

\bibitem[Moo00]{10.5555/2073946.2073993}
Andrew~W. Moore.
\newblock The anchors hierarchy: Using the triangle inequality to survive high
  dimensional data.
\newblock In {\em Proceedings of the Sixteenth Conference on Uncertainty in
  Artificial Intelligence}, UAI'00, page 397–405, San Francisco, CA, USA,
  2000. Morgan Kaufmann Publishers Inc.

\bibitem[NF16]{pmlr-v48-newling16}
James Newling and Francois Fleuret.
\newblock Fast k-means with accurate bounds.
\newblock In Maria~Florina Balcan and Kilian~Q. Weinberger, editors, {\em
  Proceedings of The 33rd International Conference on Machine Learning},
  volume~48 of {\em Proceedings of Machine Learning Research}, pages 936--944,
  New York, New York, USA, 20--22 Jun 2016. PMLR.

\bibitem[NH02]{CLARANS}
Raymond~T. Ng and Jiawei Han.
\newblock Clarans: A method for clustering objects for spatial data mining.
\newblock {\em IEEE transactions on knowledge and data engineering},
  14(5):1003--1016, 2002.

\bibitem[PCI{\etalchar{+}}07]{4270197}
James Philbin, Ondrej Chum, Michael Isard, Josef Sivic, and Andrew Zisserman.
\newblock Object retrieval with large vocabularies and fast spatial matching.
\newblock In {\em 2007 IEEE Conference on Computer Vision and Pattern
  Recognition}, pages 1--8, 2007.

\bibitem[Phi10]{philbin2010scalable}
James Philbin.
\newblock {\em Scalable object retrieval in very large image collections}.
\newblock PhD thesis, Oxford University, 2010.

\bibitem[PM99]{10.1145/312129.312248}
Dan Pelleg and Andrew Moore.
\newblock Accelerating exact k-means algorithms with geometric reasoning.
\newblock In {\em Proceedings of the Fifth ACM SIGKDD International Conference
  on Knowledge Discovery and Data Mining}, KDD '99, page 277–281, New York,
  NY, USA, 1999. Association for Computing Machinery.

\bibitem[PVG{\etalchar{+}}11]{scikit-learn}
F.~Pedregosa, G.~Varoquaux, A.~Gramfort, V.~Michel, B.~Thirion, O.~Grisel,
  M.~Blondel, P.~Prettenhofer, R.~Weiss, V.~Dubourg, J.~Vanderplas, A.~Passos,
  D.~Cournapeau, M.~Brucher, M.~Perrot, and E.~Duchesnay.
\newblock Scikit-learn: Machine learning in {P}ython.
\newblock {\em Journal of Machine Learning Research}, 12:2825--2830, 2011.

\bibitem[QPH{\etalchar{+}}10]{10.1145/1811099.1811090}
Feng Qian, Abhinav Pathak, Yu~Charlie Hu, Zhuoqing~Morley Mao, and Yinglian
  Xie.
\newblock A case for unsupervised-learning-based spam filtering.
\newblock {\em SIGMETRICS Perform. Eval. Rev.}, 38(1):367–368, jun 2010.

\bibitem[RSPR18]{rostami2018interactive}
M~Ali Rostami, Alieh Saeedi, Eric Peukert, and Erhard Rahm.
\newblock Interactive visualization of large similarity graphs and entity
  resolution clusters.
\newblock In {\em EDBT}, pages 690--693, 2018.

\bibitem[Scu10]{10.1145/1772690.1772862}
D.~Sculley.
\newblock Web-scale k-means clustering.
\newblock In {\em Proceedings of the 19th International Conference on World
  Wide Web}, WWW '10, page 1177–1178, New York, NY, USA, 2010. Association
  for Computing Machinery.

\bibitem[SR19]{FastPAM}
Erich Schubert and Peter~J Rousseeuw.
\newblock Faster k-medoids clustering: improving the pam, clara, and clarans
  algorithms.
\newblock In {\em Similarity Search and Applications: 12th International
  Conference, SISAP 2019, Newark, NJ, USA, October 2--4, 2019, Proceedings 12},
  pages 171--187. Springer, 2019.

\bibitem[SSM{\etalchar{+}}16]{10.1007/978-3-319-30303-1_12}
Mina Sheikhalishahi, Andrea Saracino, Mohamed Mejri, Nadia Tawbi, and Fabio
  Martinelli.
\newblock Fast and effective clustering of spam emails based on structural
  similarity.
\newblock In Joaquin Garcia-Alfaro, Evangelos Kranakis, and Guillaume Bonfante,
  editors, {\em Foundations and Practice of Security}, pages 195--211, Cham,
  2016. Springer International Publishing.

\bibitem[SW18]{SW18}
Christian Sohler and David~P Woodruff.
\newblock Strong coresets for k-median and subspace approximation: Goodbye
  dimension.
\newblock In {\em 2018 IEEE 59th Annual Symposium on Foundations of Computer
  Science (FOCS)}, pages 802--813. IEEE, 2018.

\bibitem[SWA{\etalchar{+}}22]{simhadri2022results}
Harsha~Vardhan Simhadri, George Williams, Martin Aum{\"u}ller, Matthijs Douze,
  Artem Babenko, Dmitry Baranchuk, Qi~Chen, Lucas Hosseini, Ravishankar
  Krishnaswamny, Gopal Srinivasa, et~al.
\newblock Results of the neurips’21 challenge on billion-scale approximate
  nearest neighbor search.
\newblock In {\em NeurIPS 2021 Competitions and Demonstrations Track}, pages
  177--189. PMLR, 2022.

\bibitem[Tal04]{T04}
Kunal Talwar.
\newblock Bypassing the embedding: Algorithms for low dimensional metrics.
\newblock In {\em Proceedings of the Thirty-Sixth Annual ACM Symposium on
  Theory of Computing}, STOC '04, page 281–290, New York, NY, USA, 2004.
  Association for Computing Machinery.

\bibitem[TZM{\etalchar{+}}20]{BanditPAM}
Mo~Tiwari, Martin~J Zhang, James Mayclin, Sebastian Thrun, Chris Piech, and
  Ilan Shomorony.
\newblock Banditpam: Almost linear time k-medoids clustering via multi-armed
  bandits.
\newblock {\em Advances in Neural Information Processing Systems},
  33:10211--10222, 2020.

\bibitem[Wai]{Waingarten}
Erik Waingarten.
\newblock Notes for algorithms for big data: Clustering.
\newblock
  \url{https://drive.google.com/file/d/1T5YYGrA3kdi4_QGvF3c_foOZvXMDaRPw/view}.

\bibitem[WWK{\etalchar{+}}12]{6248034}
Jing Wang, Jingdong Wang, Qifa Ke, Gang Zeng, and Shipeng Li.
\newblock Fast approximate k-means via cluster closures.
\newblock In {\em 2012 IEEE Conference on Computer Vision and Pattern
  Recognition}, pages 3037--3044, 2012.

\end{thebibliography}
\newpage
\appendix

\section{Additional Data}

In this section, we provide the running time data for the full range of parameters for the experiments performed in Section~\ref{sec:direct-comparison}. Table~\ref{tab:cluster-runtime-full} shows the speedups over $k$-means++, analogous to the right-hand-side table in Table~\ref{tab:runtime}. Additionally, Table~\ref{tab:cluster-runtime-absolute} provides absolute running times in milliseconds.

\begin{table}[h]
    \centering
\begin{tabular}{llrrrrrrrrr}
\toprule
          & Centers & 10   & 25   & 50   &  100  &  250  &  500  &   1000 &   2500 &   5000 \\
Dataset & Algorithm &      &      &      &       &       &       &        &        &        \\
\midrule
Census & \name  &  1.5 &  3.8 &  7.5 &  15.1 &  36.2 &  73.2 &  142.2 &  351.9 &  662.5 \\
          & \name (variance) &  0.5 &  1.1 &  2.2 &   4.6 &  11.0 &  22.2 &   43.7 &  109.5 &  214.7 \\
          & \name (covariance) &  0.2 &  0.5 &  1.1 &   2.2 &   5.2 &  10.7 &   21.0 &   54.7 &  117.4 \\
          & $k$-means++ &  1.0 &  1.0 &  1.0 &   1.0 &   1.0 &   1.0 &    1.0 &    1.0 &    1.0 \\
Song & \name  &  2.0 &  5.0 &  9.7 &  19.1 &  46.1 &  95.5 &  188.2 &  443.0 &  837.5 \\
          & \name (variance) &  0.5 &  1.1 &  2.3 &   4.5 &  11.4 &  23.1 &   44.2 &  110.9 &  217.2 \\
          & \name (covariance) &  0.2 &  0.4 &  0.8 &   1.5 &   4.0 &   8.2 &   15.5 &   40.0 &   82.4 \\
          & $k$-means++ &  1.0 &  1.0 &  1.0 &   1.0 &   1.0 &   1.0 &    1.0 &    1.0 &    1.0 \\
KDD & \name  &  1.5 &  3.7 &  6.9 &  16.3 &  39.5 &  68.3 &  158.5 &  414.7 &  727.5 \\
          & \name (variance) &  0.7 &  1.7 &  3.1 &   6.3 &  16.1 &  32.0 &   63.4 &  159.6 &  312.4 \\
          & \name (covariance) &  0.3 &  0.7 &  1.3 &   2.6 &   6.8 &  12.9 &   25.8 &   58.5 &  128.4 \\
          & $k$-means++ &  1.0 &  1.0 &  1.0 &   1.0 &   1.0 &   1.0 &    1.0 &    1.0 &    1.0 \\
Gaussian & \name  &  0.5 &  1.0 &  1.9 &   3.8 &   9.2 &  18.3 &   35.7 &   85.3 &  165.9 \\
          & \name (variance) &  0.5 &  1.0 &  2.0 &   3.8 &   9.1 &  17.7 &   34.5 &   83.6 &  162.9 \\
          & \name (covariance) &  0.4 &  1.0 &  1.7 &   3.6 &   8.2 &  16.1 &   31.8 &   79.9 &  152.6 \\
          & $k$-means++ &  1.0 &  1.0 &  1.0 &   1.0 &   1.0 &   1.0 &    1.0 &    1.0 &    1.0 \\
\bottomrule
\end{tabular}
    \caption{Average speedup when computing a clustering and assignment for different datasets relative to k-means++. In other words, each cell contains $T_{k\text{-means++}} / T_\text{PRONE}$.}
    \label{tab:cluster-runtime-full}
\end{table}

\begin{table}[h]
    \centering
    \scalebox{0.45}{
\begin{tabular}{llrrrrrrrrrrrrrrrrrr}
\toprule
          & Centers &    10   &    25   &    50   &    100  &     250  &     500  &     1000 &      2500 &      5000 \\
Dataset & Algorithm &         &         &         &         &          &          &          &           &           &         \\
\midrule
Census  &  \name  & 525.1 $\pm$ 15.3 & 534.3 $\pm$ 23.4 & 534.6 $\pm$ 20.8 & 530.1 $\pm$ 15.4 & 538.9 $\pm$ 15.2 & 533.9 $\pm$ 11.7 & 547.4 $\pm$ 23.8 & 548.2 $\pm$ 13.5 & 563.7 $\pm$ 8.7 \\
           &  \name (variance)  & 1750.6 $\pm$ 71.7 & 1778.9 $\pm$ 38.1 & 1780.3 $\pm$ 23.9 & 1752.7 $\pm$ 56.9 & 1768.6 $\pm$ 28.2 & 1757.6 $\pm$ 37.8 & 1778.9 $\pm$ 27.5 & 1761.0 $\pm$ 73.3 & 1739.6 $\pm$ 10.9 \\
           &  \name (covariance)  & 3769.0 $\pm$ 152.2 & 3882.5 $\pm$ 167.8 & 3743.9 $\pm$ 316.7 & 3714.2 $\pm$ 378.3 & 3766.4 $\pm$ 220.2 & 3662.9 $\pm$ 122.6 & 3708.7 $\pm$ 372.4 & 3525.6 $\pm$ 609.9 & 3182.2 $\pm$ 288.7 \\
           &  $k$-means++  & 812.5 $\pm$ 20.9 & 2040.6 $\pm$ 25.6 & 3994.3 $\pm$ 25.3 & 7992.9 $\pm$ 91.5 & 19519.0 $\pm$ 161.5 & 39058.4 $\pm$ 337.8 & 77823.6 $\pm$ 527.5 & 192913.1 $\pm$ 3657.0 & 373488.2 $\pm$ 221.9 \\
Song  &  \name  & 104.4 $\pm$ 5.5 & 101.8 $\pm$ 5.2 & 106.5 $\pm$ 4.1 & 109.9 $\pm$ 6.8 & 113.7 $\pm$ 8.8 & 109.0 $\pm$ 6.4 & 108.7 $\pm$ 3.9 & 117.2 $\pm$ 6.3 & 120.8 $\pm$ 12.9 \\
           &  \name (variance)  & 443.3 $\pm$ 5.0 & 448.3 $\pm$ 8.0 & 450.1 $\pm$ 9.9 & 468.0 $\pm$ 31.6 & 458.8 $\pm$ 15.1 & 450.7 $\pm$ 10.0 & 462.6 $\pm$ 11.6 & 468.3 $\pm$ 14.0 & 465.7 $\pm$ 14.9 \\
           &  \name (covariance)  & 1164.4 $\pm$ 162.1 & 1266.5 $\pm$ 136.6 & 1332.4 $\pm$ 116.3 & 1381.8 $\pm$ 108.4 & 1322.2 $\pm$ 173.4 & 1263.0 $\pm$ 175.6 & 1324.2 $\pm$ 108.4 & 1296.8 $\pm$ 177.8 & 1228.0 $\pm$ 175.2 \\
           &  $k$-means++  & 207.4 $\pm$ 4.0 & 513.7 $\pm$ 9.8 & 1036.6 $\pm$ 17.0 & 2103.9 $\pm$ 44.2 & 5245.9 $\pm$ 143.6 & 10412.5 $\pm$ 119.3 & 20464.1 $\pm$ 196.9 & 51917.4 $\pm$ 554.3 & 101149.0 $\pm$ 1962.7 \\
KDD  &  \name  & 31.2 $\pm$ 5.2 & 32.2 $\pm$ 7.4 & 34.1 $\pm$ 1.8 & 28.9 $\pm$ 4.9 & 29.2 $\pm$ 9.1 & 34.6 $\pm$ 6.3 & 30.8 $\pm$ 8.6 & 28.8 $\pm$ 5.2 & 33.4 $\pm$ 6.6 \\
           &  \name (variance)  & 72.0 $\pm$ 1.4 & 71.8 $\pm$ 1.4 & 76.5 $\pm$ 7.2 & 75.1 $\pm$ 7.1 & 71.8 $\pm$ 1.2 & 73.9 $\pm$ 2.1 & 77.0 $\pm$ 4.6 & 74.7 $\pm$ 3.2 & 77.7 $\pm$ 5.9 \\
           &  \name (covariance)  & 169.3 $\pm$ 2.9 & 173.5 $\pm$ 5.5 & 180.3 $\pm$ 13.1 & 184.1 $\pm$ 21.2 & 170.8 $\pm$ 8.9 & 182.6 $\pm$ 19.0 & 188.7 $\pm$ 16.9 & 204.1 $\pm$ 40.3 & 189.1 $\pm$ 11.6 \\
           &  $k$-means++  & 48.0 $\pm$ 0.3 & 119.3 $\pm$ 3.7 & 233.9 $\pm$ 3.2 & 470.4 $\pm$ 11.6 & 1153.5 $\pm$ 2.1 & 2363.3 $\pm$ 108.6 & 4878.1 $\pm$ 171.1 & 11928.9 $\pm$ 234.3 & 24285.9 $\pm$ 243.8 \\
Gaussian  &  \name  & 30.0 $\pm$ 2.3 & 29.1 $\pm$ 2.2 & 30.6 $\pm$ 3.3 & 29.1 $\pm$ 1.4 & 29.1 $\pm$ 1.0 & 28.7 $\pm$ 0.2 & 29.3 $\pm$ 0.4 & 30.2 $\pm$ 0.6 & 31.5 $\pm$ 0.6 \\
           &  \name (variance)  & 29.8 $\pm$ 3.0 & 28.9 $\pm$ 0.3 & 29.5 $\pm$ 1.0 & 29.6 $\pm$ 0.6 & 29.6 $\pm$ 0.3 & 29.6 $\pm$ 0.4 & 30.2 $\pm$ 0.3 & 30.8 $\pm$ 0.2 & 32.1 $\pm$ 0.6 \\
           &  \name (covariance)  & 31.8 $\pm$ 2.2 & 31.5 $\pm$ 2.1 & 34.6 $\pm$ 6.6 & 30.8 $\pm$ 0.3 & 32.7 $\pm$ 2.8 & 32.6 $\pm$ 2.4 & 32.8 $\pm$ 2.3 & 32.2 $\pm$ 0.5 & 34.3 $\pm$ 0.4 \\
           &  $k$-means++  & 13.7 $\pm$ 2.2 & 30.0 $\pm$ 0.4 & 59.5 $\pm$ 2.2 & 111.5 $\pm$ 2.8 & 268.6 $\pm$ 1.5 & 525.3 $\pm$ 1.9 & 1043.3 $\pm$ 1.4 & 2575.3 $\pm$ 5.0 & 5229.6 $\pm$ 135.0 \\
\bottomrule
\end{tabular}
}

    \caption{Average running time and standard deviation in milliseconds when computing a clustering and assignment for different datasets relative to k-means++.}
    \label{tab:cluster-runtime-absolute}
\end{table}
\end{document}